\documentclass[sigconf]{acmart}

\usepackage{booktabs} 
\usepackage{multirow}
\usepackage{xspace}
\usepackage{enumitem}
\usepackage{array}
\usepackage[font=small,labelfont=bf,tableposition=top]{caption}
\DeclareCaptionLabelFormat{andtable}{#1~#2  \&  \tablename~\thetable}

\copyrightyear{2020}
\acmYear{2020}
\setcopyright{acmcopyright}\acmConference[CIKM '20]{Proceedings of the 29th ACM International Conference on Information and Knowledge Management}{October 19--23, 2020}{Virtual Event, Ireland}
\acmBooktitle{Proceedings of the 29th ACM International Conference on Information and Knowledge Management (CIKM '20), October 19--23, 2020, Virtual Event, Ireland}
\acmPrice{15.00}
\acmDOI{10.1145/3340531.3411906}
\acmISBN{978-1-4503-6859-9/20/10}

\begin{document}
\newcommand{\ig}[1]{\textcolor{blue}{$\ll$\textsf{#1 --IG}$\gg$}}
\newcommand{\dnote}[1]{\textcolor{red}{$\ll$\textsf{#1 --DS}$\gg$}}
\newcommand{\mnote}[1]{\textcolor{magenta}{$\ll$\textsf{#1 --MD}$\gg$}} 
\newcommand{\dt}[1]{\textcolor{purple}{$\ll$\textsf{#1 --DT}$\gg$}}
\newcommand{\an}[1]{\textcolor{brown}{$\ll$\textsf{#1 --AN}$\gg$}}
\newcommand{\ad}[1]{\textcolor{yellow}{$\ll$\textsf{#1 --AD}$\gg$}}

\newcommand{\ecom}{{e-commerce}\xspace}
\newcommand{\remove}[1]{}

\def\w#1{\texttt{\small #1}}

\newcolumntype{L}[1]{>{\raggedright\let\newline\\\arraybackslash\hspace{0pt}}m{#1}}
\newcolumntype{C}[1]{>{\centering\let\newline\\\arraybackslash\hspace{0pt}}m{#1}}
\newcolumntype{R}[1]{>{\raggedleft\let\newline\\\arraybackslash\hspace{0pt}}m{#1}}

\fancyhead{}
\title{E-Commerce Dispute Resolution Prediction}
\author{David Tsurel}
\affiliation{
  \institution{The Hebrew University of Jerusalem}
}
\email{dmtsurel@mail.huji.ac.il}

\author{Michael	Doron}
\affiliation{
  \institution{The Hebrew University of Jerusalem}
}
\email{michael.doron@mail.huji.ac.il}
		
\author{Alexander Nus}
\affiliation{
  \institution{eBay Research}
}
\email{alnus@ebay.com}
	
\author{Arnon Dagan}
\affiliation{
  \institution{eBay Research}
}
\email{ardagan@ebay.com}

\author{Ido	Guy}
\affiliation{
  \institution{eBay Research}
}
\email{idoguy@acm.org}

\author{Dafna Shahaf}
\affiliation{
  \institution{The Hebrew University of Jerusalem}
}
\email{dshahaf@cs.huji.ac.il}

\begin{abstract}
E-Commerce marketplaces support millions of daily transactions, and some disagreements between buyers and sellers are unavoidable. Resolving disputes in an accurate, fast, and fair manner is of great importance for maintaining a trustworthy platform. Simple cases can be automated, but intricate cases are not sufficiently addressed by hard-coded rules, and therefore most disputes are currently resolved by people. In this work we take a first step towards automatically assisting human agents in dispute resolution at scale. We construct a large dataset of disputes from the eBay online marketplace, and identify several interesting behavioral and linguistic patterns. We then train classifiers to predict dispute outcomes with high accuracy. We explore the model and the dataset, reporting interesting correlations, important features, and insights.




\end{abstract}

%
%



\maketitle
\section{Introduction}
\label{sec:intro}

The connection between sellers and buyers is at the core of online marketplaces such as eBay or Amazon~\cite{recsys}. These large e-commerce marketplaces see millions of daily transactions, and therefore some conflicts inevitably occur, from an unreceived package to a product being different than expected. Many disputes are resolved by direct communication between buyer and seller, yet not always an agreement can be reached. In such cases, the dispute has to be resolved by the marketplace platform, typically by applying a human arbitrator to examine and resolve cases. This kind of decision making is essential for an online marketplace to take care of the interests of both sides and establish user trust. Yet, as the number of transactions grows, manual arbitrator work becomes a burden.


Automating the arbitration process is of great importance, and it is common practice to use simple automated rules such as ``if tracking information shows that the item has not arrived, and the seller does not respond, the buyer wins the case''. As arbitrators follow very specific guidelines in their arbitration workflow, it would seem that an automatic rule-based system would be sufficient. 

However, many cases require a broader outlook, which is hard to capture with a rule-based system. Misunderstandings or missing information can be resolved by examining buyer and seller data and the correspondence between them. For example, in cases where there is a fraud concern, an arbitrator will want to inspect the history of both the buyer and the seller for previous suspicious behavior. In addition, arbitration requires understanding natural language used by both sides to fully comprehend their claims. Textual messages are especially useful to fill gaps in other signals, like wrong shipping or tracking information.

We propose to aid human arbitrators in their decision making with a model predicting the final resolution of the dispute. We gathered a large dispute dataset consisting of claim features, transaction features, seller features, buyer features, and textual communication features. Seller and buyer features include 
past behavior on the site, demographic features, general priors, and priors related to the transaction in dispute. By using this data to predict outcome, we hope to give human arbitrators a first approximation of the final result. This could help save manpower and cope with the fast-growing amount of transactions (and consequently, disagreements).


To the best of our knowledge, this is the first comprehensive study of dispute analysis and automatic resolution in \ecom. 
%
Our contributions in this paper are: 
\begin{itemize}
\item We collect and analyze a dataset of disputes between buyers and sellers in an e-commerce platform. We explore the dataset, reporting interesting correlations and properties 
that we discovered through the exploration (Section \ref{sec:dataset}).
\item We train models for predicting dispute outcome (Section \ref{sec:predict}).
We develop a classifier that reaches AUC of $0.94$ with precision of $89\%$ and recall of $88\%$. 

\item We describe the model results and perform ablation studies to assess the importance of features and feature families, and find that integrating various aspects of the data is crucial for performance, as no single feature family suffices for accurate classification. We analyze and characterize errors made by the model (Section \ref{sec:results}). 
\item We add an interpretability module to our model, which assists humans in understanding the reasoning behind the predicted decision of a specific case. It includes a feature importance component explaining the contribution of different features to prediction, as well as a component for textual feature interpretation that highlights predictive tokens. 

\item We analyze the effect disputes have on users (both during and after the dispute). In particular, we saw that losing a dispute has a negative effect on the number of transactions made after the dispute ended, and that dispute outcome is reflected in politeness strategies used during correspondence.
\end{itemize}

\section{Bigger Picture}


In today's world, people are frequently subjected to predictive algorithms. Such algorithms are increasingly used to make important decisions affecting human lives, ranging from approving financial loans to social welfare benefits. 


As courts are overwhelmed with the sheer volume of cases \cite{steelman1997have}, judges are now guided by algorithms in a growing number of state courts.
These algorithms mostly focus on determining a defendant's risk, bail decisions, sentencing length, recidivism and parole \cite{monahan2016risk,brennan2009evaluating}. One widely used criminal risk assessment tool is COMPAS (Correctional Offender Management Profiling for Alternative Sanctions). COMPAS has been used to assess more than one million offenders since 1998 \cite{dressel2018accuracy}. COMPAS uses 137 features about an individual, including past criminal record.
Although these ``algorithm-in-the-loop'' studies provide tools to assist human agents in legal decision making, they keep the human agent in charge as the final arbiter. This is partially due to algorithmic limitations, and partially due to the desire to keep the normative role of judges in the hands of human decision making, as discussed by Morison and Harkens \cite{Morison}. Of note is a study by Sela \cite{sela} who showed that participants experience more procedural justice when the final arbitrator is human. Our work likewise attempts to provide informed resolutions, and not take the decision away from the final human decision maker.

\paragraph{Fairness and Bias}

Recently there have been growing concerns about the use of such algorithms and their \emph{fairness} \cite{dressel2018accuracy}. For example, although these algorithms are not allowed to use race as an input, an analysis revealed that the predictions were racially biased, and black defendants are substantially more likely to be classified as high risk \cite{angwin2016machine}. 
The issue of fairness is a serious one. Despite the fact that these tools are meant to support decisions, not make them, research has shown that when people receive specific advisory guidelines they tend to follow them in lieu of their own judgment \cite{fourpr,rossi2019building,gunning2017explainable}. 
In the case of judges, it is also somewhat risky for them to release someone contrary to AI's recommendation; in private correspondence, one judge expressed the sentiment that ``no one wants to find themselves on the front page of the newspapers, if that person were to commit another crime''. 

Our goal in this work is to take first steps towards building a similar system for e-commerce dispute resolution. To mitigate bias, we add an \emph{interpretability} component, helping agents understand the reasoning behind predictions. However, we acknowledge that this topic needs to be further investigated in future work. 

\vspace{-1mm}

\paragraph{Interpretability}

In an effort to tackle these issues, many algorithms incorporate interpretability components that help shed light on their recommendations and possible biases. Interpretabiliity allows models to provide explanations for why different decisions and predictions were made, based on the features provided in training and prediction time \cite{kim}. Many interpretability methods were recently studied \cite{molnar}, among them SHAP \cite{NIPS2017_7062} -- a method to assign importance values to features, and LIME \cite{ribeiro2016should} -- a method that learns a local approximation of the classification, explaining which features influenced its decision. These tools enable black-box machine learning models to be more transparent, thus hopefully preventing undesired bias from influencing the decision making process.

\section{Related Work}
\label{sec:rew}


In this work, we focus on the problem of \emph{online disputes in e-commerce}. To the best of our knowledge, this is the first work that provides an automatic tool for prediction of this problem. However, while automating the final resolution of the Online Dispute Resolution (ODR) was not handled in e-commerce, several attempts to automate the process were done in the parallel field of legal intelligence (e.g., predicting judicial decisions) \cite{lawlor,Zeleznikow,Kleinberg}. 

%

Dispute resolution in \ecom has been a challenge since inception. An early attempt by legal experts to resolve disputes in eBay was conducted by a non-binding mediator, and a formal set of rules was not yet established \cite{katsh1999commerce}.
Since the emergence of ODR, tools have been built to assist human arbitrators and participants in dispute resolution. 
A review by Goodman \cite{goodman} found that automated ODR systems were able to handle more participants, increasing revenue to their companies. Early works were based on combining defined rules and knowledge databases that could consult participants in a dispute \cite{Carneiro}. Xu et al \cite{xu2012analyzing} used Latent Dirichlet Allocation on eBay Motors dispute data to predict whether participants would reach a settlement, by training a conversation topic model and comparing agreement level between the topic distribution in each participant's messages. However, none of these works developed a fully-automated dispute resolution system.

Recently, Zhou et al \cite{zhou2019legal} predicted the results of lawsuits that followed unresolved e-commerce disputes by using data taken from those disputes. In contrast to our work on predicting dispute outcomes \emph{within} the e-commerce platform, they focus on the outcome of an external legal process that follows customer dissatisfaction from the ODR process. As they state, only a small minority of the buyers choose to engage in such a perplexing and expensive process~\cite{zhou2019legal}, and the data set is therefore small and biased. 

Several studies tried to learn the dynamics of ODRs without predicting their result. Wang et al \cite{wang} used sentiment analysis to study the question of whether or not a Wikipedia discussion would escalate into a dispute. Friedman et al \cite{friedman} studied how anger can help or harm one's case in an online dispute. 

These tools, while providing assistance and auxiliary information to human agents, do not tackle the direct problem of predicting the result of the dispute, which is the main task of human arbitrators in e-commerce ODR systems. In contrast, our work predicts the outcome of disputes, which could help human agents reach a decision faster. Another important difference is that our work provides interpretable features that explain the reasons for the resolution, helping human arbitrators make informed decisions. 










\remove{
\dnote{to check: Re-engineering justice? Robot judges, computerised courts and (semi) automated legal decision-making, Do Judges Need to Be Human? The Implications of Technology for Responsive Judging, Predicting judicial decisions of the European Court of Human Rights: A natural language processing perspective, Project dispute prediction by hybrid machine learning techniques, Inductive learning of dispute scenarios for online resolution of customer complaints, Optimizing parameters of support vector machine using fast messy genetic algorithm for dispute classification, Machine learning for intelligent support of conflict resolution, Improving classification accuracy of project dispute resolution using hybrid artificial intelligence and support vector machine models, eMediation-Towards Smart Online Dispute Resolution.}

\dnote{copy-paste from https://dl.acm.org/doi/pdf/10.1145/3359152}

The rise of machine learning has fundamentally altered decision making: rather than being made solely bypeople, many important decisions are now made through an “algorithm-in-the-loop” process where machinelearning models inform people

Society lacks both clear normative principles regarding howpeople should collaborate with algorithms as well as robust empirical evidence about how people do collaboratewith algorithms. Given research suggesting that people struggle to interpret machine learning models and toincorporate them into their decisions—sometimes leading these models to produce unexpected outcomes—it isessential to consider how different ways of presenting models and structuring human-algorithm interactionsaffect the quality and type of decisions made

For although calls to adopt machine learning modelsoften focus on the accuracy of these tools [14,46,59,66], accuracy is not only attribute of ethicaland responsible decision making. The principle of procedural justice, for instance, requires thatdecisions be (among other things) accurate, fair, consistent, correctable, and ethical [55]. Even asalgorithms bear the potential to improve predictive accuracy, their inability to reason reexivelyand adapt to novel or marginal circumstances makes them poorly suited to achieving many ofthese principles [2]. As a result, institutions implementing algorithmic advice may find themselveshailing the algorithm's potential to provide valuable information while simultaneously cautioningthat the algorithm should not actually determine the decision that is made [74]

\dnote{copy-paste from https://dl.acm.org/doi/pdf/10.1145/3287560.3287590}
Humans are the final decision makers in critical tasks that involveethical and legal concerns, ranging from recidivism prediction, tomedical diagnosis, to fighting against fake news

However, it is important to recognize different roles that machinelearning can play in different tasks in the context of human decisionmaking. In tasks such as object recognition, human performancecan be considered as the upper bound, and machine learning modelsare designed to emulate the human ability to recognize objects in animage. A high accuracy in such tasks presents great opportunitiesfor large-scale automation and consequently improving our soci-ety's efficiency. In contrast, efficiency is a lesser concern in taskssuch as bail decisions. In fact, full automation is often not desiredin these tasks due to ethical and legal concerns. These tasks arechallenging for humans and for machines, but with vast amounts ofdata, machines can sometimes identify patterns that are unsalient,unknown, or counterintuitive to humans. If the patterns embeddedin the machine learning models can be elucidated for humans, theycan provide valuable support when humans make decisions

\dnote{copy paste from https://www.cs.cornell.edu/home/kleinber/aer18-fairness.pdf}

One can easily
imagine how this could happen since recidivism
predictions will be polluted by the fact that past
arrests themselves may be racially biased. In
fact, a recent ProPublica investigation argued
that the risk tool used in one Florida county
was in fact discriminatory (Angwin et al. 2016). 

\dnote{copy-paste from http://proceedings.mlr.press/v81/chouldechova18a/chouldechova18a.pdf -- which we might cite}

However, the use of predictive analytics in the
area of child welfare is contentious. There is
the possibility that some communities—such as
those in poverty or from particular racial or ethnic groups—will be disadvantaged by the reliance
on government administrative data because they
will typically have more data kept about them
simply by dint of being poor and on welfare. Such
families could then be flagged as high risk and
be more frequently investigated. If the algorithm
uses past investigations to produce a high risk
score for a family, then this will exacerbate the
original bias.
On the other hand, these analytics tools can
augment or replace human judgments, which
themselves are potentially biased. There is a possibility that caseworkers are basing their screening decisions in part on personal experiences or
current caseloads. Caseworker decisions may
also be be affected by cognitive biases—for example over-weighting recent cases, unrelated to
the current case, where a child has been fatally
harmed. When making decisions under time
pressure, caseworkers might be guilty of statistical discrimination, where they use easily observed
features (e.g. living in a neighborhood with high
crime rates) as proxies for unobservable but more
pertinent attributes (e.g. drug-use). Bias in human decision-making is often difficult to assess,
and the existing research does not provide a con

Some of the earliest research comparing human predictions to
those of statistical models goes back to the pioneering work of Meehl (1954). Decades of research and several large scale meta-analyses have
largely upheld the original conclusions: When it
comes to prediction tasks, statistical models are
generally significantly more accurate than human
experts (Dawes et al., 1989; Grove et al., 2000;
Kleinberg et al., 2017).

\dnote{the below is copy-paste from SIGIR! change change change}

Legal Intelligence. Judicial decision prediction is not a novel topic
which has been raised since 1960s that an article appeared in The
American Behavioral Scientist entitled “Using Simple Calculations
Predict Judicial Decisions [28]”. Since then a number of researchers
from legal field started to explore the possibilities and methodologies to approach such problem [16, 23, 29, 42]. For instance, some
advocates claimed that the computers can help find and analyze
the law as well as helping lawyers and lower court judges to predict or anticipate the judicial decision [23]. Correspondingly, some
methodologies were adopted to predict the probability of a favorable decision experimented on specific types of cases [16, 29]. In
the recent work, Oard et al. studied the information retrieval for
e-discovery [31, 32]. Wang et al. [40] proposed a model for crime
classification. Despite those opposition voices who were skeptical to
the judgments made by machine instead of human [42], the advantages of objectivity and justice brought by the automatic judgment
prediction should not be ignored [11]. However, their approaches
never have a chance to really implement, especially for e-commerce
ecosystem, because of data barrier/sparseness, algorithm limitation, and lack of computational legal knowledge. In this study, we
pioneer this problem by using multiview dispute data along with
sophisticated multi-task learning and deep dispute representation
learning

Online Dispute Resolution in E-Commerce. 20 years ago,
the scholars have predicted the growth of online disputes while
e-commerce was becoming an increasingly important place for
transactions. There is reason to believe that dispute resolution
systems and services are needed to be online [20]. With the development of e-commerce as well as techniques, Online dispute
resolution (ODR) system has become mature nowadays which is a
form of online settlement that uses alternative methods for dispute
resolution. Today almost all the e-commerce platforms operate on
their own ODR systems. Though ODR can be done in a way of low
cost and high efficiency, there is no recommendation provided for
the customers who are not satisfied with the resolution result and
about to file a lawsuit. In this work, we leverage the large amount
of dispute data provided by the e-commerce platform to enable
lawsuit judgment prediction. On the other hand, legal judgment
prediction is also a way of making the resolution of e-commerce
disputes more legitimate [12]

}

\section{Problem Formulation}

The eBay \emph{Resolution Center} is meant to help both buyers and sellers in case they have a problem with an item they bought, or sold, on eBay. With over 60 million disputes per year, it is one of the biggest ODR systems in the world~\cite{rule2016designing}. The two most common reasons for opening disputes are not receiving an item (Item not recieved, INR), or receiving an item that is significantly not as described in the original listing (SNAD). Sellers can be either professional businesses (B2C) or private individuals (C2C).

When opening a dispute, both buyers and sellers are advised to contact the other side (seller or buyer, respectively) before reporting an issue, and see if they can work things out. 
Once escalating an issue in the Resolution Center, both sides have a period of a few days 
to reach an agreement. 
If the issue is not resolved, the dispute is moved to the resolution of an arbitrator on behalf of eBay. During the entire period, the seller and buyer can communicate via messages, which will later be available to the arbitrator to be considered for resolution purposes. 

A transaction can be escalated more than once, for example when one of the sides wishes to appeal a previously made resolution, or when the system decides to reopen a case. The escalating party can 
therefore change over the course of the dispute. 

The arbitrator has access to a variety of signals, from those related to the case itself (delivery receipt, tracking number, price, etc.), through activity history of both buyer and seller, to the message correspondence between the buyer and seller about the case. In some cases, these signals take a long time to process, so human arbitrators can resolve only a handful per hour. Our conversations with several team leads of regional dispute resolution centers indicated that the current process is tedious and involves many technical details. ``The agreement between human arbitrators would be high, so each case is assigned with just one arbitrator, but this is still a lot of work'' said one of them and another noted that ``\textit{receiving automatic assistance in this task can be of big help to our team.''}






The human arbitrator follows a decision-tree guide, with some of the nodes leading to a deterministic decision criterion.
The arbitrator follows the decision criteria in each node in the decision tree until they reach a resolution. In many cases this process is sufficient for case resolution, and is relatively straightforward. This process can be automated, and indeed simple cases have been automated. For example, one case involved a user complaining that their package has not been received. The following is the conversation between the buyer and the seller.

\begin{itemize}[leftmargin=*]
    \item Buyer: ``I have not received this item. Where is it?''

    \item Seller: ``Your order was sent out to you.  Please don't worry, we have checked with the shipping company.  They told us the parcel is delayed some days but now could arrive at your customs.  Please wait  one more week for releasing. If you don't receive it, please email us, we will provide you an emergent solution.''
\end{itemize}

\remove{
\begin{itemize}
    \item Buyer: ``I haven't received it yet. Would you have the shipping status? Thank you''

    \item Seller: ``Dear friend  Thanks for your contact,please don't worry,we shipped the package to you with tracking number: [Redacted] tracking information shows the item has arrive at your Country, so could  you please contact your local post to check the issue first?  Any questions,please feel free email us  We are here to help  Best Regards''

    \item Buyer: ``Oh, Great!  I'll be waiting for this.  thanks buddy''   
    \item Seller: ``It is our pleasure ,any idea,please feel free let us know  Thank you  Best Regards''   
\end{itemize}
}

The dispute reached the resolution center, which checked the tracking number and concluded that ``tracking shows no movement and the buyer has not confirmed receipt''. This is a simple case since tracking shows the buyer has not received the item, and the dispute was therefore automatically resolved in favor of the buyer.  

However, not all cases are so simple. Several reasons prevent an automatic rule-based process from being an effective dispute resolution system. The first is the need to {\bf integrate several aspects of the case} to get a broader outlook, which is hard to achieve with a rule-based system. 

For example, a buyer purchased a Blackberry cell phone in an online auction for the price of \$142.99. The buyer sent the seller several messages claiming they received an empty box, and then opened a dispute. Here are some of the messages sent by the buyer:

\begin{itemize}[leftmargin=*]
	\item ``hi friend!    today i send a payment for this cell phone curve 9360''
	\item ``hello dear friend,    the purchase of this cell phone has been a mess, I received a empty box, I have been very sad about this problem because I bought this cell phone to my mother, I need you to help me with a refund of the money, please      I'll give you positive feedback and 5 stars''
	\item ``hello dear friend!    I sent pictures of the empty box that I received''
\end{itemize}

The seller has had a long tenure (over 12 years), is a high-volume seller, and has been involved in over 2,000 disputes with other buyers. When this dispute reached a human arbitrator they ruled against the buyer, with the following explanation: ``[Buyer] has 8 cases open as an empty box received and they are all similar items ... we can no longer cover this buyer due to their fraud risk''. By examining the buyer's history of ordering phones and claiming that the boxes were empty, the dispute resolver managed to detect fraud. 
This case is more complicated, as it required the arbitrator to examine not just the transaction details, but also the previous behavior of the users involved. Automating this type of dispute will be more challenging, since simple rules would be too broad to capture the details of different cases. 


An additional challenge for automation is processing natural language used by the buyer and seller to gain better understanding of their claims and assertions. In another case, a user complained that the purchased item was not as described, since it was too small.

\begin{itemize}[leftmargin=*]

	\item Buyer: ``Hello. I emailed you before the purchase of this bag and you didn't reply. Now this bag was delivered today and it is not as you stated LARGE.... Which I originally questioned you about at first. I did contact eBay about this situation because I will not keep a small bag because it's too small for me
	'' 

	\item Buyer: ``Hello you have this bag said as large but when calling a store that carry these bags stated that those measurements are considered to be medium. Pls explain.''

	\item Seller: ``I've told you all I can. I'm not a store. I'm just a seller.  I have nothing more to say.''

	\item Buyer: ``You don't have too and I respect that because I'm a child of God and He fights all of my battles, so with that being said I have been informed to let eBay handle this, god bless.''

	\item Seller: ``Sounds good to me.''

\end{itemize}

The dispute reached the resolution center that ruled in favor of the seller. The arbitrator had to use the messages exchanged between the buyer and the seller to determine that the seller provided exact measurements in the product listing, and therefore was not responsible for the buyer's assumptions about the product's size. In this case analyzing transaction features or user history was not enough, and understanding user text was also necessary.

We see that different cases require examining and understanding all of the available data sources to determine the correct outcome.

In this work, we focus on {\bf automating the dispute resolution process}, where the ground truth is the resolution as decided by expert human arbitrators. In particular, we train a classifier to learn from past cases and, by integrating different aspects of the case, predict the resolution of disputes between buyers and sellers. 




\remove{
\begin{figure}
    \begin{center}
        (A)
    \end{center}

    \begin{itemize}
    \item Buyer: ``I haven't received it yet. Would you have the shipping status? Thank you''

    \item Seller: ``Dear friend  Thanks for your contact,please don't worry,we shipped the package to you with tracking number: [Redacted] tracking information shows the item has arrive at your Country, so could  you please contact your local post to check the issue first?  Any questions,please feel free email us  We are here to help  Best Regards''

    \item Buyer: ``Oh, Great!  I'll be waiting for this.  thanks buddy''   

    \item Seller: ``It is our pleasure ,any idea,please feel free let us know  Thank you  Best Regards''   
\end{itemize}

\begin{center}
\rule{0.7\linewidth}{0.4pt}

\smallskip

        (B)
    \end{center}

\begin{itemize}
	\item ``hi friend!    today i send a payment for this cell phone curve 9360''
	\item ``hello dear friend,    the purchase of this cell phone has been a mess, I received a empty box, I have been very sad about this problem because I bought this cell phone to my mother, I need you to help me with a refund of the money, please      I'll give you positive feedback and 5 stars''
	\item ``hello dear friend!    I sent pictures of the empty box that I received''
	\item ``hello dear friend,    I am very sad because I have received an empty package of this curve 9360 cell phone that I buy from you, please I need your help because I buy this cell phone to give to my mother and was stolen by someone, I want you help me to send me a replacement or a refund send money please.    I'll give you positive feedback and 5 stars''
\end{itemize}

\begin{center}
\rule{0.7\linewidth}{0.4pt}

\smallskip

        (C)
    \end{center}

\begin{itemize}

	\item Buyer: ``Hello. I emailed you before the purchase of this bag and you didn't reply. Now this bag was delivered today and it is not as you stated LARGE.... Which I originally questioned you about at first. I did contact eBay about this situation because I will not keep a small bag because it's too small for me, sorry....PLEASE CONTACT ME BACK ASAP.'' 

	\item Buyer: ``Hello you have this bag said as large but when calling a store that carry these bags stated that those measurements are considered to be medium. Pls explain.''

	\item Seller: ``I've told you all I can. I'm not a store. I'm just a seller.  I have nothing more to say.''

	\item Buyer: ``You don't have too and I respect that because I'm a child of God and He fights all of my battles, so with that being said I have been informed to let eBay handle this, god bless.''

	\item Seller: ``Sounds good to me.''
\end{itemize}


    \caption{Caption}
    \label{fig:my_label}
\end{figure}
}

\section{Dataset and Characteristics}
\label{sec:dataset}

In our research, we constructed a dataset of online disputes. In this section we report some of its attributes, and observations gathered during analysis. 


\subsection{Dispute Dataset}
Our data\footnote{We describe the dataset in detail, both quantitatively and qualitatively, but cannot publicly share it due to the sensitivity of the data. 
} consists of disputes that occurred on eBay between 2010-2019 with buyers using the US version of the website, pertaining to products from 8,354 categories. We sampled 1,000,000 messages exchanged between buyers and sellers, filtered out messages that were duplicated due to table joins, and aggregated the messages into 72,023 buyer-seller conversations.

While there are over 40 resolution options (like partial refunds, timeout resolutions, item arrival during the dispute, third party fault, etc.), in this paper we focus on cases with two clear cut resolutions - when the arbitrator
actively ruled in favor of either the seller or the buyer. The distribution of these labels was 42,880 for seller wins (59.6\%) and 29,143 for buyer wins (40.4\%). This rules out other cases, such as those where one of the sides withdrew their complaint, those where the two sides reached a settlement without the need for arbitration, or those where the ruling was technical (e.g., for lack of response by one of the sides).

\remove{
\begin{figure}[t]
\centering
\includegraphics[width=0.98\linewidth]{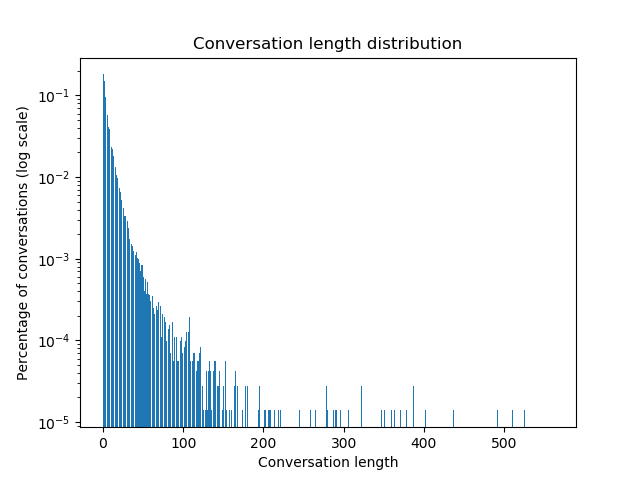}
\caption{\small Distribution of conversation length between buyer and seller (logarithmic scale). Most of the conversations are rather short, with a median conversation of just 4 messages.
\label{fig:conversation_distribution}}
\end{figure}
}

\subsection{Features and Feature Families}

The data had 937 features, which belong to several feature families: 
\begin{itemize}[leftmargin=*]
  \item Claim - features related to the claim (e.g., type of claim, which party escalated first)
  \item Transaction - feature related to the transaction before the claim (e.g., price of the item)
  \item Claim seller - claim features related to the seller (e.g., seller tenure days, b2b or c2c)
  \item Claim buyer - claim features related to the buyer (e.g., number of disputes buyer participated in the last year). We analyzed the demographics of buyers that were involved in disputes. Most users did not specify their gender, but of those who did 71.5\% identified as male, compared to just 28.5\% who identified as female. 
    79\% of buyers in our dataset, limited to buyers browsing 
    the US version of the site, are themselves from the US, and the rest are from other countries (2\% from Russia, 1.3\% from Israel, 1.3\% from Brazil, and the rest of the countries have less than 1\%).

  \item Seller data - features related to the seller user profile (e.g., city of residence, currency, number of email accounts)
  \item Buyer data - features related to the buyer user profile (e.g., tax status, anonymous email)
  \item Textual features -
    One of the key features we examined is the conversation between the disputing buyer and seller in its different stages: before a purchase is made, before a dispute is opened, and during the dispute. Most conversations are short: the median conversation is just $4$ messages 
    long, but there is a long tail and some conversations can reach hundreds of messages,
    and this long tail skews the average (8.6) and the standard deviation (17.9).
    
    The language of the conversation can have several registers. Buyers usually use everyday vernacular language that can reach acrimonious tones and insults if the dispute gets heated. Sellers, especially those who are professional businesses, use a combination of free-form 
    language and predefined templates (e.g., ``Dear Customer: Your payment has been received. The order will be shipped out today. Shipping time needs about 20-25 business days to arrive at your address ... Thanks for your purchase Best Regards'').
    Their tone is generally appeasing, but can also become harsh if they are upset.

    We processed the conversations that transpired between the buyer and the seller. The messages where standardized using case folding, stemming, and stop word removal. We trained a fastText classifier\cite{bojanowski2017enriching,joulin2016bag} on an independent dataset of 1,000,000 messages and their dispute outcomes. This classifier was then used to generate textual features for disputes in our dataset - both the outcome and the embeddings of this fastText classifier were used as features.  
\end{itemize}

\remove{
Studying the feature correlations in the dataset allowed us to understand the inner structure of the system better. \dnote{this paragraph is really sad. does it look like we gained any better understanding of the system to you?}
For example, low-price items often resulted in an INR (item not received) dispute, possibly due to sellers choosing less reliable shipping methods for low-price items.
%
Another correlation was between B2C (Business-to-Costumer) sellers, longer tenure, more disputes and more responsiveness to buyers' claims than C2C (Costumer-to-Costumer) sellers. This joint feature set might represent the more professional attitudes of businesses who sell in higher volumes than private individuals.



}

\section{Predicting Dispute Outcome}
\label{sec:predict}
Our main task is to predict the outcome of disputes in the dataset. In this section, we describe the basic feature correlations with the output, present the classifiers we trained for the classification task, and explain the hyperparameter optimization process. Given a dispute, our task is to predict whether the buyer wins or the seller wins, based on the features of the case. We define ``seller wins'' as the positive class for classification purposes.

\subsection{Correlations with Outcome}
We first checked the correlation between each feature and the dispute outcome. The results are presented in table \ref{corr_table}. We can observe that important features come from several feature families, including claim features (like first escalating party), user features (like seller history), and textual features (like fastText prediction and embedding). This is an initial indication that different aspects of the dispute have an effect on the outcome, and that combining different aspects could be beneficial to prediction.


\begin{table}[t]
\caption{Correlation (absolute value) between dispute outcome and other features.}
\label{corr_table}
\begin{tabular}{@{}llc@{}}
\toprule
Feature & Feature Family & |Correlation|\\ 
\midrule
First escalating party                   & Claim & 0.54                            \\ 
Recent escalating party                  & Claim & 0.51                            \\ 
Seller info last modified date            & Seller data    & 0.38                            \\ 
Seller credit card on file            & Seller data & 0.37                            \\ 
fastText prediction                    & Textual  & 0.35                            \\ 
Claim type (INR/SNAD)                & Claim & 0.32                            \\ 
\bottomrule
\end{tabular}
\vspace{-2mm}
\end{table}

Using KL-divergence~\cite{carmel12kl}, we also examined \emph{words} (unigrams and bigrams) that appear more frequently in cases where the buyer wins the dispute and words that appear more frequently when the seller wins. 


SNAD (significantly not as described) claims pertain to disputes where the buyer claims the item is drastically different than the description in the e-commerce platform. Accordingly, the unigrams and bigrams were related to attributes of the item itself (\w{dresses}, \w{the} \w{size}, \w{retro}, etc.).


INR (item not received) claims pertain to disputes where a buyer claims the item was not received at all. Accordingly, unigrams and bigrams were related to residence attributes and delivery situations (e.g., \w{apartment}, \w{porch}, \w{valid} \w{tracking}, \w{distribution} \w{center}). An interesting property of textual features in INR cases is that features indicating seller wins often assign responsibility to a third party (e.g., \w{neighbors}, \w{porch}, \w{mailman}, \w{mailbox}, \w{stolen}, \w{my} \w{door}, \w{old} \w{address}, etc.), while features indicating a buyer win often describe a problem in the shipping process (\w{cds} -- corporate delivery service, \w{valid} \w{tracking}, \w{fedex}, \w{been} \w{weeks}, etc.).





\subsection{Classifiers}
Our goal is to build a classifier for automatically predicting the outcomes of disputes in the dataset. We tested and evaluated several classifier families to find a model that achieves the best performance. We used the scikit-learn implementations for all of the classifiers~\cite{scikit-learn}.
Each classifier was tested by averaging the results of a 5-fold cross-validation. We examined the following classifiers: 

\begin{itemize}[leftmargin=*]
    \item Majority - a simple baseline classifier that always predicts the same label, the most frequent label in the dataset (which happens to be ``seller wins'', with 59.6\% of the resolutions).
    \item Gaussian Naive Bayes - a classifier that assumes features are independent, and each feature is normally distributed.
    \item K-nearest-neighbors - a classifier that predicts based on the labels of the closest neighbors in the feature space.
    \item Decision Tree - a tree-structured classifier where leaves represent labels and internal nodes split based on values of a given feature.
    \item Random Forest - an ensemble model of decision trees that uses bootstrapping to improve accuracy and lower over-fitting.
    \item Gradient Boosted Trees (XGBoost) - an ensemble model of decision trees that uses gradient descent to introduce new trees that improve upon the error of previous trees.
    \item Neural Network - a network of feedforward layers of neurons used to predict labels.
\end{itemize}

We examined several metrics to evaluate the performance of the proposed models: accuracy, precision, recall, F1 score (harmonic mean of precision and recall), and area under the ROC curve (henceforth AUROC). AUROC was our main metric for comparing classifiers, as it captures a classifier's ability to distinguish between different outcome classes.

In addition to testing these classifiers on the full dataset, we also tried to segment the dataset into subsets based on two prominent features that split the dataset into different scenarios: these are claim type (SNAD or INR) and seller type (B2C or C2C). For each classifier type, we trained distinct classifiers for each segment, and averaged the results. We then examined the outcome to see if segmenting the datasets into different scenarios helped achieve better performance.

\subsection{Hyperparameter Optimization}
\begin{table}[t]
\caption{Performance of different classifiers on predicting dispute outcomes. Precision and recall are calculated for ``seller wins'' predictions.}
\label{classifier_results_table}
\setlength{\tabcolsep}{0.17em}
\begin{tabular}{@{}lC{1.05cm}C{1.3cm}C{1.2cm}C{1cm}C{1cm}@{}}
\toprule
& AUROC & Accuracy & Precision & Recall & F1 \\ \midrule
Majority   & 0.5                      & 0.60             & 0.60              & \textbf{1.0}             & 0.75              \\ 
KNN & 0.60                    & 0.60             & 0.65              & 0.71           & 0.68              \\ 
Neural Network       & 0.62                    & 0.61             & 0.64              & 0.80           & 0.71              \\ 
Na\"ive Bayes       & 0.72                    & 0.65             & 0.72              & 0.73           & 0.69              \\ 
Decision Tree    & 0.90                    & 0.83             & 0.86              & 0.86           & 0.86              \\ 
Random Forest     & 0.92                    & 0.84             & 0.85              & 0.89           & 0.87              \\ 
XGBoost         & \textbf{0.94}                    & \textbf{0.86}             & \textbf{0.89}              & 0.88           & \textbf{0.89}              
\\ 
\bottomrule
\end{tabular}
\vspace{-4mm}
\end{table}

To optimize our classifiers, we used hyperparameter tuning with the objective of maximizing AUROC on an independent validation dataset of 70,671 disputes, generated in the same way as described in section \ref{sec:dataset}.
As exhaustive grid search can be computationally prohibitive, we used a randomized search technique. Each classifier was evaluated on a 5-fold split validation dataset for 50 possible hyperparameter configurations. Some classifiers did not have hyperparameters to optimize: Majority and Gaussian Naive Bayes. The following hyperparameter spaces were optimized for the rest of the classifiers:
\begin{itemize}[leftmargin=*]
    \item K-nearest-neighbors (KNN) - number of neighbors (1-10), weights (uniform or distance-based), distance metric (Manhattan or Euclidean). The optimal hyperparamers were found to be 7 closest neighbors, distance-based weights, Euclidean distance.
    \item Decision Tree - maximum depth (10-number of features), minimum number of samples to split an internal node (2-20), minimum number of samples to split a leaf (2-20). The optimal hyperparameters were found to be maximum depth of 101, 10 minimum number of samples to split an internal node, and 20 to split a leaf.
    \item Random Forest - number of trees used (10-200), maximum depth (10-number of features), minimum number of samples to split an internal node (2-20), minimum number of samples to split a leaf (2-20). The optimal hyperparameters were found to be 178 trees, maximum depth of 72, 11 minimum number of samples to split an internal node, and 12 to split a leaf.
    \item Gradient Boosted Trees (XGBoost) - number of trees (150-1000), learning rate (0.01-0.6), maximum depth (10-number of features), subsample (0.3-0.9), column subsample (0.5-0.9), minimum child weight (1-4). The optimal hyperparameters were found to be 225 trees, 0.025 learning rate, maximum depth of 309, 0.55 subsample, 0.67 column subsample, minimum child weight of 3.
    \item Neural Network - number of hidden layers (1-3), neurons in every layer (32, 64, 128, 256), activation function (tanh, relu, logistic), solver (adam or lbfgs), L2 regularization parameter (0.0001, 0.001, 0.01). The optimal hyperparameters were found to be a single hidden layer of 256 neurons, tanh activation, lbfgs solver, and 0.001 L2 regularization.
\end{itemize}


\section{Results}
\label{sec:results}

In this section, we report the results of the classifiers presented in the previous section. We then go into further analysis of the best-performing model and its interaction with the dataset by examining feature importance and feature ablation. We augment the model with an interpretability module that enhances its transparency, and analyse the model errors. We conclude by studying the effect of disputes on user behavior.











\begin{figure}[t]
\centering
\includegraphics[width=0.98\linewidth]{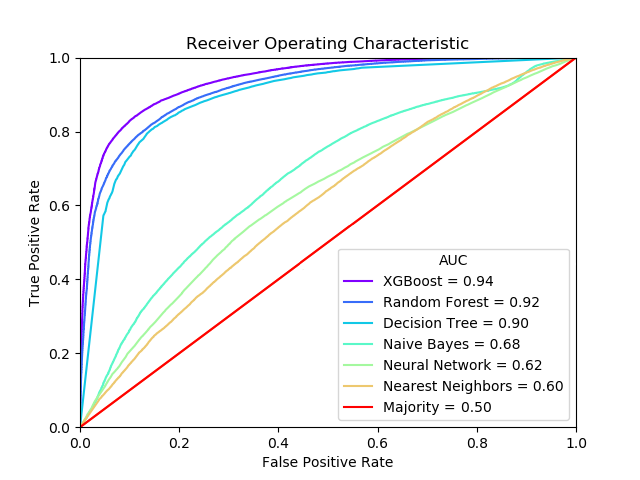}
\caption{\small ROC curves of the different classifiers. \label{fig:roc_curves}}
\end{figure}





\subsection{Classification Results}
We ran the classifiers described in the previous section on our dataset. The results are presented in Table \ref{classifier_results_table}. We can see that the XGBoost classifier achieved better results than other classifiers in most metrics (except recall, which was trivially dominated by the Majority classifier), including an AUROC value of 0.94. the Random Forest and Decision Tree classifiers also yielded high performance. 

The ROC curves are presented in Figure \ref{fig:roc_curves}. We can see the ``majority'' classifier has no discriminative power as it does not observe the data points, and it therefore lies on the diagonal. Other classifiers have higher predictive power, with XGBoost reaching 0.94. 

As discussed in the previous section, we also attempted to segment the dataset based on claim type (SNAD or INR) and seller type (B2C or C2C). However, when we evaluated classifiers on the different segments, we did not observe an improvement in the measured metrics. For example, AUROC of XGBoost (weighted average of segments) was only 0.92 
compared to 0.94 on the full dataset. As we will see in the next sections, claim type and seller type were important features, but were not the most important. It seems that segmenting the dataset into subsets lowered the predictive power of classifiers due to having less data, and the advantage of segmentation into different scenarios was not enough to compensate.

We note that for deployment of such systems in practice, it is crucial to have access to as much information as the arbitrator has.  



\subsection{Feature Importance}

To better understand our dataset and model, we examined feature importance in the XGBoost classifier. 

First, we examined XGBoost gain (Table \ref{xgboost_gains_table}), which is the average contribution of a feature across all splits where it is used, and compared it to feature correlations with the seller winning the dispute (as previously presented in Table \ref{corr_table}). Using the correlation with the output, we can gain quick insight regarding the average effect of the feature on the identity of the winner. Interestingly, although the feature whose gain is highest is also most correlated with the dispute outcome (``First escalating party''), the next top gain feature (``Has seller responded to claim?'') is the 192nd most correlated feature with the outcome, hinting on a more complex relationship between the features and the outcome.

Importantly, the top features in terms of gain are not dominated by one feature family. Some are related to the claim (which party escalated, claim type (INR/SNAD), whether the seller responded, etc.), and others are features of the seller and buyer (Is the seller top rated, seller and buyer countries, etc.). Interestingly, textual features were not found to have high gain.

\begin{table}[t]
\caption{Gain of top features in the XGBoost model. The correlation (absolute value) of each feature with the outcome is also presented.}
\label{xgboost_gains_table}
\renewcommand{\arraystretch}{1.1}
\begin{tabular}{@{}L{3.1cm}lcc@{}}
\toprule

Feature       & Feature Family  & Gain & |Correlation| \\ 
\midrule
First escalating party     & Claim                   & 119.65      &   0.54       \\ 
\midrule[0.03em]
Has seller responded to claim?      & Claim                   & 39.99      & 0.14           \\ 
\midrule[0.03em]
Recent escalating party     & Claim                      & 26.32   & 0.51              \\
\midrule[0.03em]
Claim type (INR/SNAD)   & Claim                          & 17.13           & 0.32      \\
\midrule[0.03em]
Seller site locale     & Seller data           & 13.30       & 0.23          \\
\midrule[0.03em]
Seller information last modified date  & Seller data                            & 11.85       & 0.38          \\
\midrule[0.03em]
Seller country     & Seller data                     & 11.78        & 0.19         \\
\midrule[0.03em]
Is seller account confirmed?      & Seller data                   & 11.19        & 0.06         \\
\midrule[0.03em]
Is seller top-rated?      & Seller data                 & 10.32      & 0.30            \\
\midrule[0.03em]
Has seller responded to claim before escalation?   & Claim  & 7.84   &  0.14             \\
\bottomrule
\end{tabular}
\end{table}

\subsection{Feature Ablation Study}
\begin{figure}[t]
\centering
\includegraphics[width=0.98\linewidth]{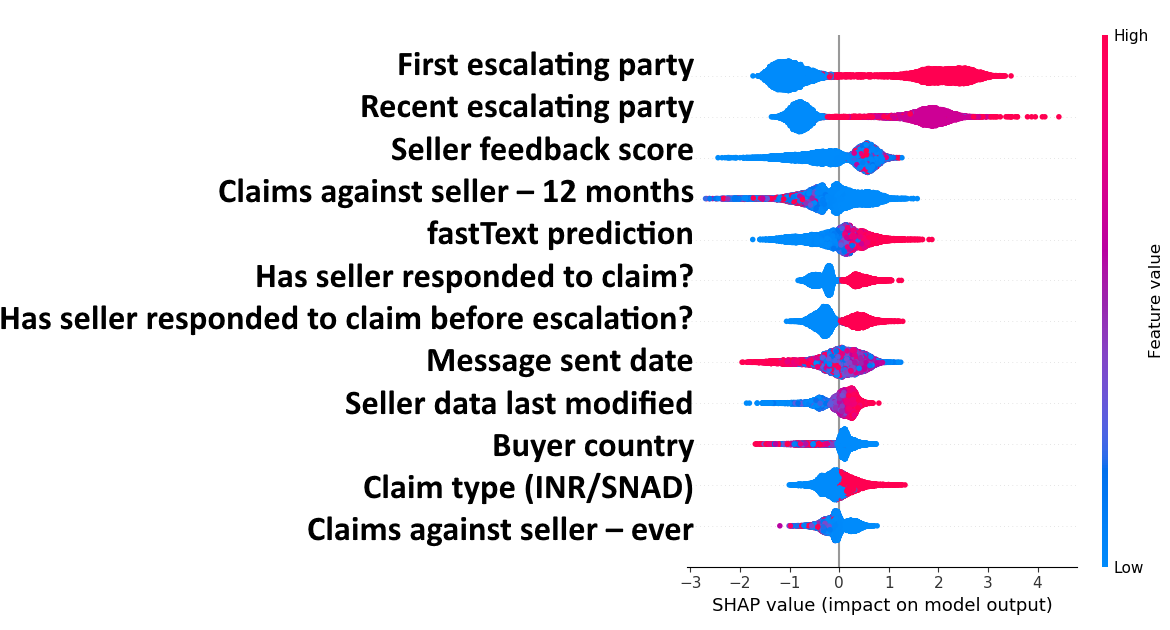}
\caption{\small SHAP value importance of top features. The features are presented in descending order of SHAP value impact. Each dot represents an instance from the test set (red for high values, blue for low values), and its location on the horizontal axis represents the effect of that value on the model prediction.
\label{fig:shapley_values}}
\end{figure}

\begin{table}[t]
\caption{Performance of the XGBoost model when trained on a single feature family.}
\label{feature_families_ablation}
\setlength{\tabcolsep}{0.37em}
\renewcommand{\arraystretch}{1.1}
\begin{tabular}{@{}lC{1.05cm}C{1.3cm}C{1.2cm}C{1cm}C{1cm}@{}}
\toprule
  & AUROC & Accuracy & Precision & Recall & F1 \\ 
\midrule
All Features                & 0.94                     & 0.87              & 0.89               & 0.89            & 0.89               \\ 
Claim              & 0.84                     & 0.77              & 0.81               & 0.79            & 0.80                \\ 
Transaction        & 0.62                     & 0.61              & 0.64               & 0.81            & 0.71               \\ 
Claim seller       & 0.79                     & 0.75              & 0.76               & 0.86            & 0.80                \\ 
Claim buyer        & 0.59                     & 0.61              & 0.64               & 0.78            & 0.70                \\ 
Seller data        & 0.82                     & 0.76              & 0.78               & 0.83            & 0.80                \\ 
Buyer data & 0.63                     & 0.63              & 0.66               & 0.79            & 0.72               \\ 
Textual           & 0.70                      & 0.67              & 0.70                & 0.79            & 0.74               \\ 
All purchase            & 0.85                     & 0.78              & 0.82               & 0.80             & 0.81               \\ 
All buyer                  & 0.64                     & 0.63              & 0.66               & 0.80             & 0.72               \\ 
All seller                 & 0.85                     & 0.78              & 0.79               & 0.87            & 0.83               \\ 
All user                   & 0.87                     & 0.80               & 0.81               & 0.89            & 0.84               \\ 
\bottomrule
\end{tabular}
\vspace{-2mm}
\end{table}

Due to the computational problem of enumerating all possible feature splits, XGBoost uses a greedy algorithm for choosing features by their relative gain \cite{chen2016xgboost}. To portray a more accurate picture of the contribution of each feature to the final model, we also conducted a series of ablation tests. We measured feature importance using SHAP values. SHAP values use a game-theoretic approach to find which features deserve the most credit by measuring the loss generated by removing that feature from the model, over all possible permutations\cite{NIPS2017_7062}. The features with the highest impact on the outcome for our XGBoost model are presented in Figure \ref{fig:shapley_values}. The analysis shows highly contributing features to be what we could expect in a dispute, including important claim features (such as claim type (INR/SNAD), escalating party, and whether the seller responded to the claim), textual features related to the conversation between the two parties, as well as other features related to the history and credibility of the parties (including seller feedback score, how many disputes they were involved in, and more).

\remove{
\begin{figure*}
    \centering
    \begin{minipage}{0.47\textwidth}
        \centering
        \includegraphics[width=1.02\linewidth]{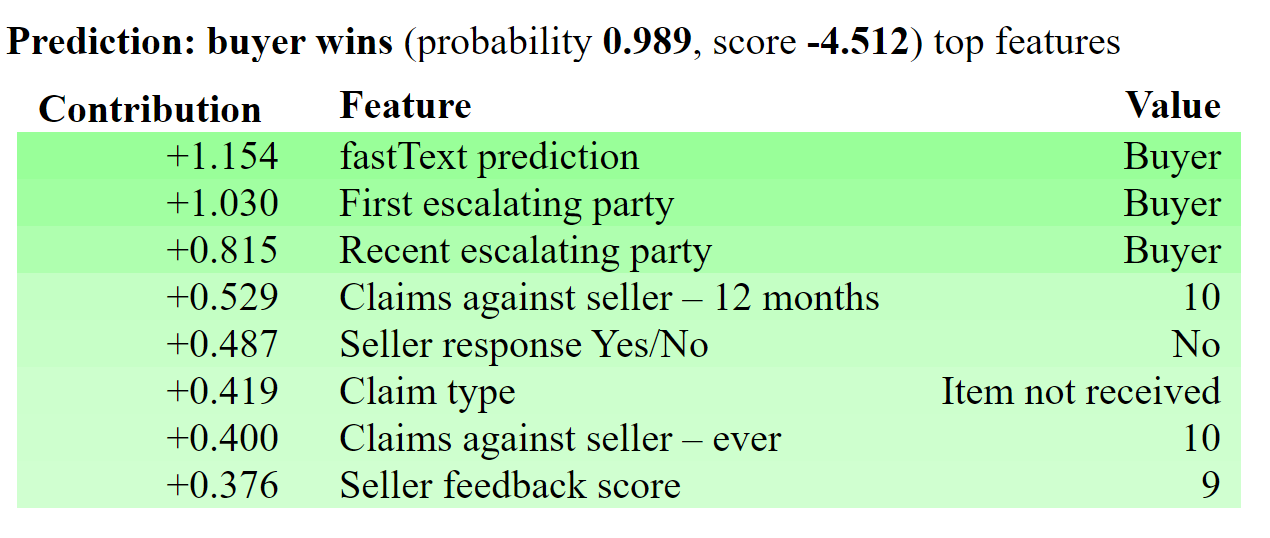}
        \caption{\small An example of an interpretation for a decision made by a classifier in a specific case. The top contributing features are shown, along with the value of the feature and its contribution to the decision. 
        \label{fig:explainable}}
    \end{minipage}
    \hfill
    \begin{minipage}{0.47\textwidth}
        \centering
        \includegraphics[width=1\textwidth]{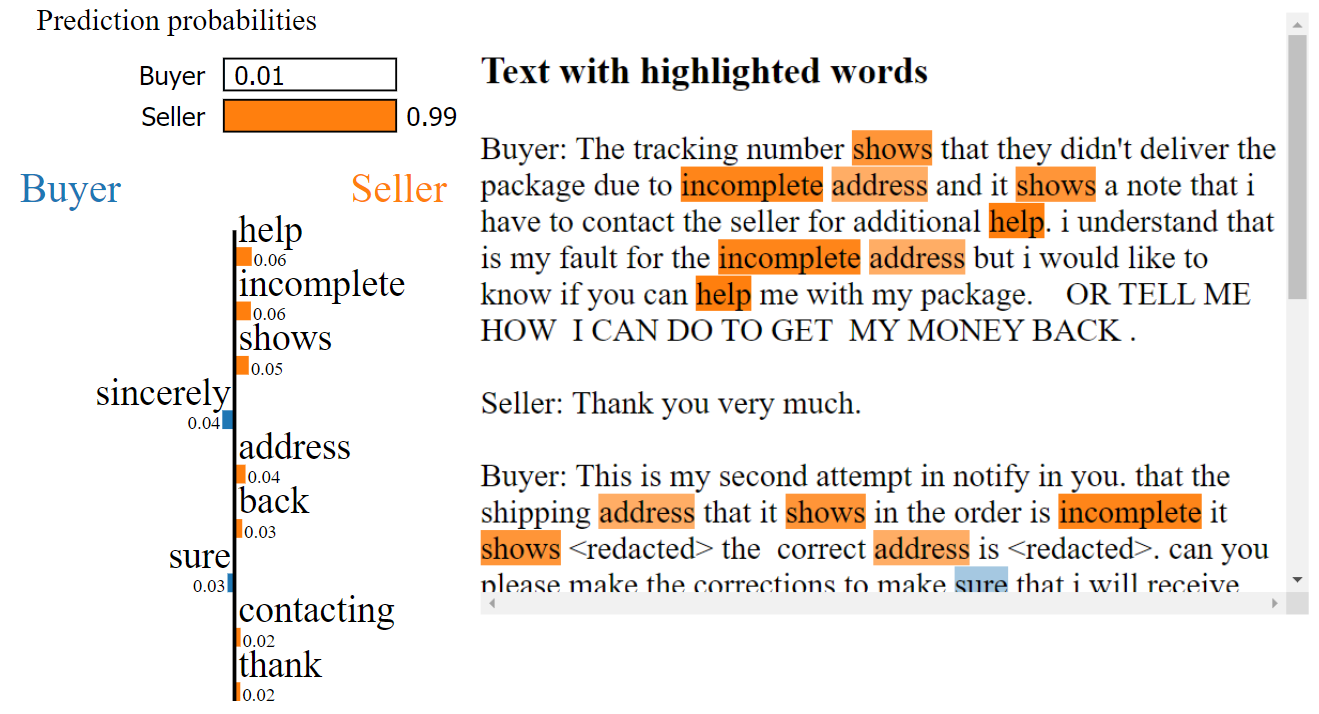} 
        \caption{\small An example of text analysis using LIME. On the top left we see the model's prediction, on the bottom left the contribution of important tokens, and on the right the text with highlighted tokens.
        \label{fig:LIME_text}}
    \end{minipage}
\vspace{-2mm}
\end{figure*}
}

\begin{figure}[t]
\centering
\includegraphics[width=1.02\linewidth]{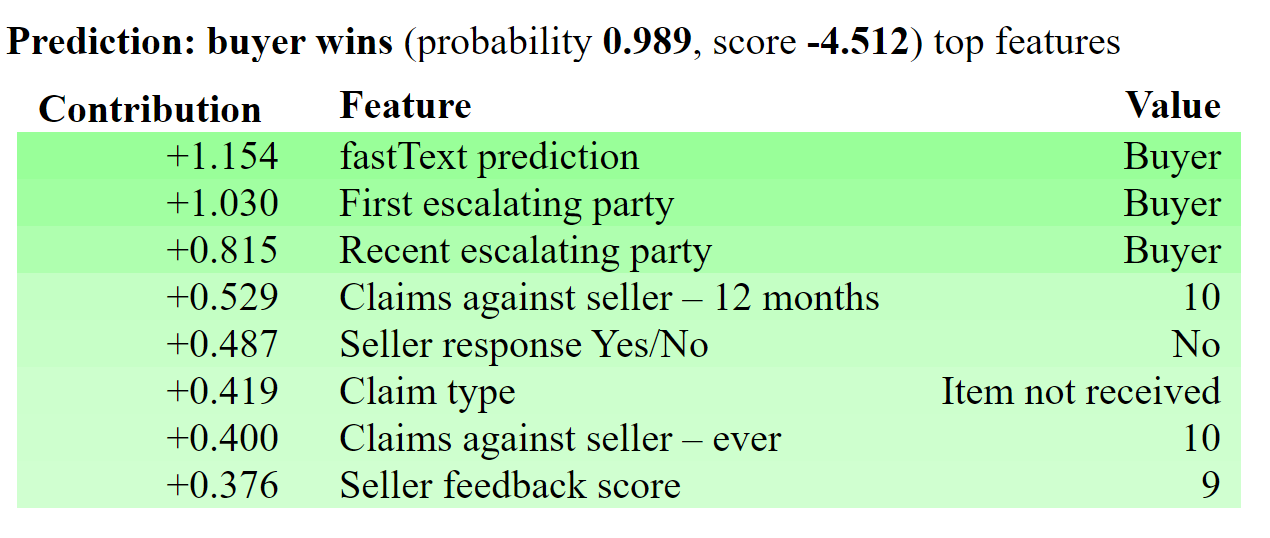}
\caption{\small An example interpretation for an automatic decision. The top contributing features are shown, along with the value of the feature and its contribution to the decision. 
\label{fig:explainable}}
\vspace{-3mm}
\end{figure}


\begin{figure}[t]
\centering
\includegraphics[width=1\linewidth]{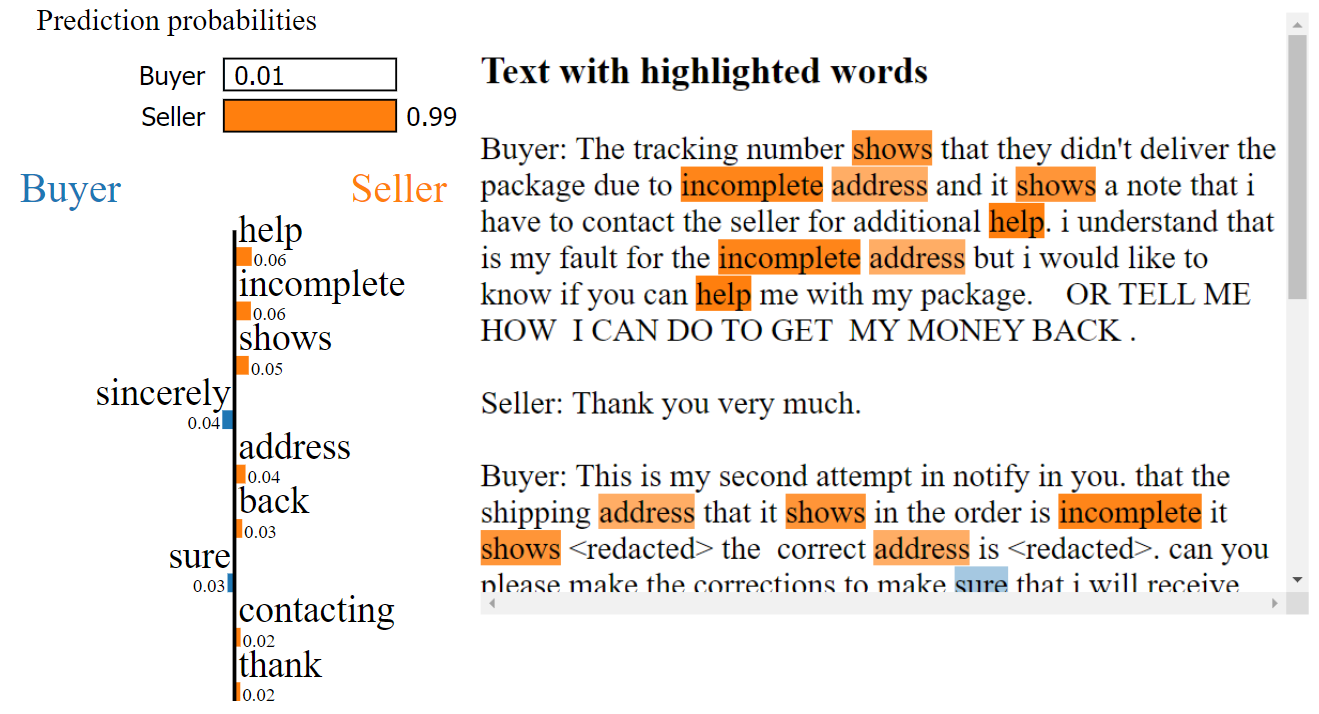} 
\caption{\small An example of text analysis using LIME. On the top left we see the model's prediction, on the bottom left the contribution of important tokens, and on the right the text with highlighted tokens.
\label{fig:LIME_text}}
\vspace{-5mm}
\end{figure}

We furthered our ablation tests by explicitly training the model with all the features except one. Model performance never dropped below 0.94 when ablating any single feature. This shows that our dataset is \emph{robust}, and no single feature is dominant enough that the model could not reach good results without it.

We also examined the feature contribution by training the model with only a single feature at a time. We were able to reach as much as 0.79 AUROC when using the ``first escalating party'', significantly less than the 0.94 AUROC of the full model. This testifies to the non-triviality of the problem, as no single feature is dominant enough to correctly classify the whole dataset.

Finally, we examined contributions of features families, 
as listed in Section~\ref{sec:dataset}: claim features, transaction features, claim seller features, claim buyer features, seller data features, buyer data features, and textual features. We also examined combined feature families such as all purchase features (combining transaction and claim), all seller features, all buyer features, and all user features (including both buyer and seller). Results are presented in Table \ref{feature_families_ablation}, showing that no single feature family succeeded in reaching high AUROC, and a combination of several families was necessary. Features regarding seller data reached the highest AUROC value of 0.85, while the lowest AUROC (0.59) was reached when using only buyer features. 
It is possible that sellers have higher impact on disputed situations -- by aspects such as quality of manufacturing, proper shipping methods, and accurate item descriptions. It is also possible that sellers have a more consistent behavior than buyers, and that the large volume of sales on the seller side is reflected in more accurate representation in the data. Indeed, sellers in our dataset were involved in 11 times as many transactions as buyers, on average.

The claim feature family, which had top gain features such as claim type (INR/SNAD) and which party escalated, reached 0.84 AUROC. Even combined feature families such as ``all user features'' reached only 0.87 AUROC. We see that \emph{no single feature family} is enough to accurately classify the dataset, and that observing and integrating various aspects of the data achieves better performance.

\subsection{Prediction Interpretability}

\remove{
\begin{figure}[t]
\centering
\includegraphics[width=0.98\linewidth]{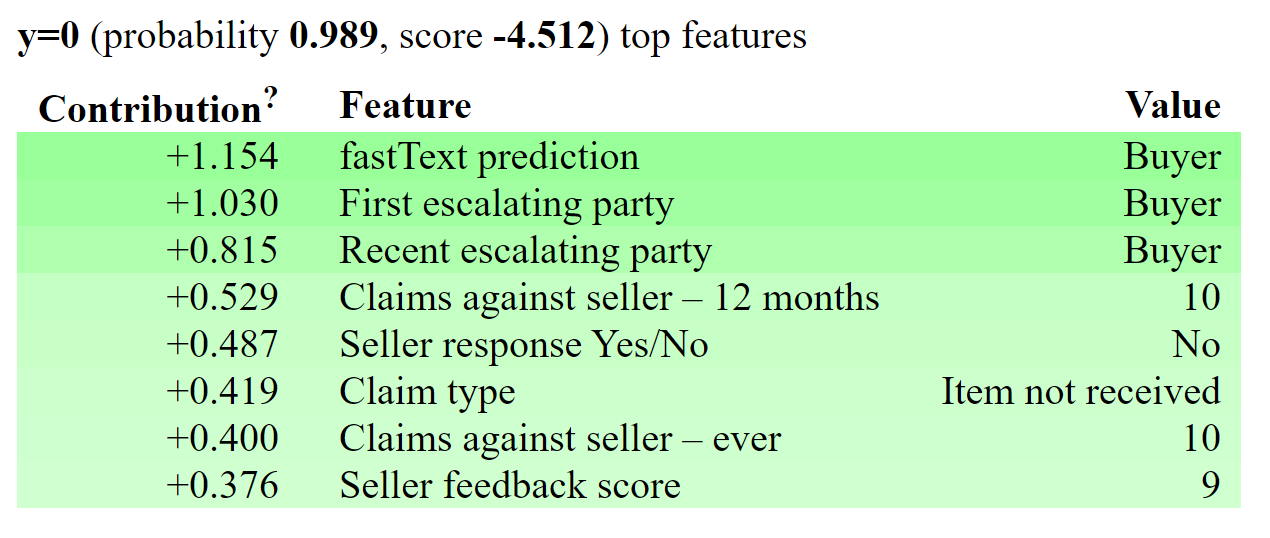}
\caption{\small explainable \label{fig:error_analysis_eli5}}
\end{figure}
}
Fairness and transparency are important in AI decision-making in general, and especially in online dispute resolution, where the arbitrator needs to make fair and informed decisions. 
For this reason, we added interpretability capabilities to our model, so that a human arbitrator can quickly understand how the model reached its decision. 

To explain the decision of our model for a specific case, we used the ELI5 tree explainer for XGBoost~\cite{ELI5}. For a specific decision, it captures the contribution of each feature to the decision by the paths followed within trees in the ensemble. It sums the contribution of each node in the path, which indicates how much the overall score was changed by going from the parent node to its child. An example explanation is shown in Figure~\ref{fig:explainable}, where the top contributing features are shown. In this case, the buyer claimed the item was not received and escalated the dispute, while the seller did not reply to the buyer's claim and was involved in numerous disputes in the past (10) with a relatively low feedback score (9). The most important feature in this case was the fastText classification of user correspondence. In the correspondence, the buyer repeatedly asks the seller when their purchase will be sent, with no response from the seller. In such a case, our classifier decided to rule in favor of the buyer. We can see that using the interpretability tool can aid a human arbitrator by succinctly pointing out features that are important for the specific case, and explaining the reasoning behind the classifier's predicted decision.

\noindent \textbf{Interpreting textual features.} FastText prediction is often selected as a top-contributing feature. However, informing an agent that the text was important for the decision is not very insightful. Thus, we further used LIME~\cite{ribeiro2016should} to interpret the textual features gathered from conversations between the disputing sides. LIME learns a local approximation around the prediction, which enables it to assign feature importance for classifiers even when such a task is not straightforward, such as the neural network embedding generated by fastText. We use LIME to highlight the tokens that most affected the outcome of the fastText classifier, which makes textual features interpretable, and also allows a human arbitrator to quickly focus on important terms in the conversation.

We present an example in Figure \ref{fig:LIME_text}. In this case, the tokens ``incomplete'' and ``address'' were deemed important, and the phrase ``incomplete address'' is highlighted several times in the text. The buyer concedes that they have entered an incomplete address and it is their own fault, but still asks the seller for help. The fastText classifier predicts that this case will be ruled in favor of the seller, and this is indeed the same decision made by the human arbitrator. Notice that the phrase ``tracking number'', which would have been important in many other INR cases, is not highlighted here - in this case it is not as important. 
This kind of tools could provide the arbitrator with more transparency into both structured and textual information used by the classifier to reach its decision, which can help a human-in-the-loop become more effective and trustworthy.


\subsection{Error Analysis}

\begin{figure}[t]
\centering
\includegraphics[width=0.98\linewidth]{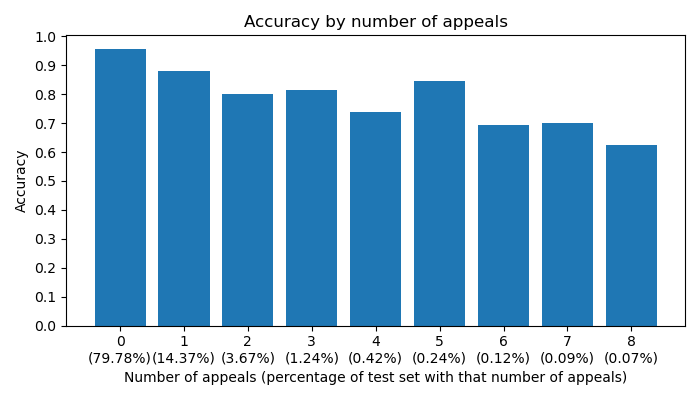}
\caption{\small Model accuracy on the test set, grouped by number of appeals. In cases with no appeals (79.78\% of cases), our model had a high accuracy rate of 0.96. Accuracy was lower in cases with more appeals, dropping to 0.88 when there was one appeal (14.37\% of cases), and to 0.8 when there were two appeals (3.67\% of cases).  
\label{fig:error_analysis_appeals_length}}
\vspace{-2mm}
\end{figure}


Our model reached high AUROC (0.94). In this section, we examined the characteristics of cases where the model still failed. One observation we made was that model accuracy decreased in cases where one of the sides appealed, with accuracy inverse to the number of appeals in the case. From 0.96 accuracy in cases with no appeal, down to 0.88 in cases with one appeal and 0.8 in cases with two appeals. This might stem from the fact that cases with appeals tend to be generally harder to decide and the ground truth is not obvious, with both sides displaying substantial argumentation. Indeed, in many of these cases, the initial decision was reversed by a second human agent, sometimes after being presented with new information about the case. An indirect indicator of this was the length of agent decision summary -- incorrectly classified disputes had significantly longer summaries (923 characters on average versus 618), indicating appeals or complex cases. Note that this effect of agent summary length disappeared when we only looked at the length of the first summary or summaries without appeals, meaning that the length difference is due to the appeals themselves. 

Another interesting aspect was the frequency of certain words in the correctly- and incorrectly- classified disputes. First, the word \w{appeal} appeared in incorrectly classified disputes 200\% more than in correctly classified disputes, fitting our result above. Incorrectly classified disputes had more words indicating communication between the agent and the buyer/seller (\w{contact}, \w{educate}, \w{@}, \w{email}) compared to correctly classified disputes (20\%, 33\%, 80\% and 173\% more, respectively). Note that \w{educate} here is a reserved word, used when an agent teaches the seller/buyer regarding the procedures.     

\remove{
\subsection{Error Analysis}
Despite reaching 0.94 AUROC, our model still reached an inaccurate prediction in several cases. We examined some of these cases, and found that several instances had intricacies that complicated resolutions in these cases.

One case where the item was not received went through a round of appeals by both sides over the question of who was responsible for providing an incorrect address. The buyer and seller provided claims and evidence directly to the arbitrator (some of it by phone call), critical information that was not available to our classifier, and which caused the overturning of the decision by a second arbitrator. In another case, the seller initially provided an incorrect tracking number and therefore lost the case. The seller then appealed and provided the correct tracking number, and the dispute resolution was changed. These cases shows the need for an appeal process, and also that our classifier will not necessarily work when essential information is missing.
}


\subsection{How Disputes Affect Users}

\begin{figure}[t!]
\centering
\includegraphics[width=0.98\linewidth]{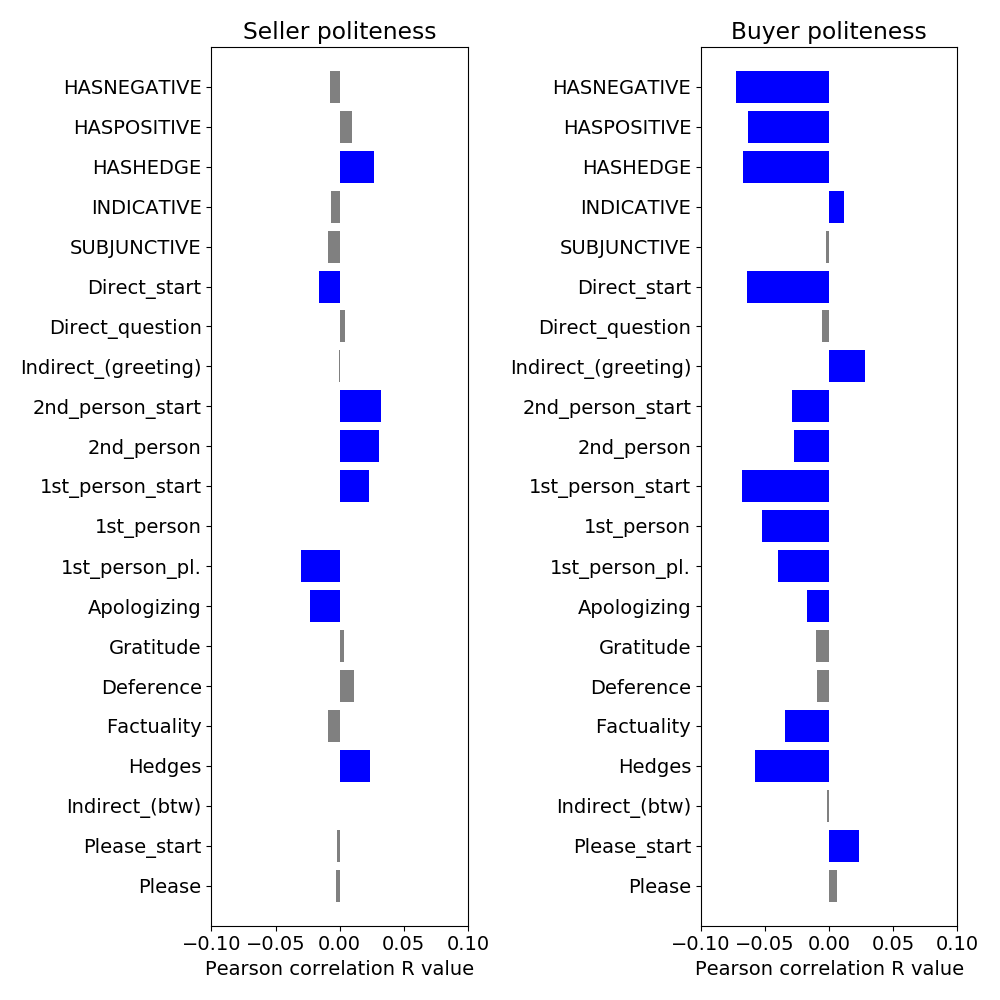}
\caption{\small Correlation of politeness strategy in first dispute message with that participant winning. Blue is significant (p < 0.005).
\label{fig:first_polite}}
\vspace{-3mm}
\end{figure}

\begin{figure*}[t!]
\centering
\includegraphics[width=1\linewidth]{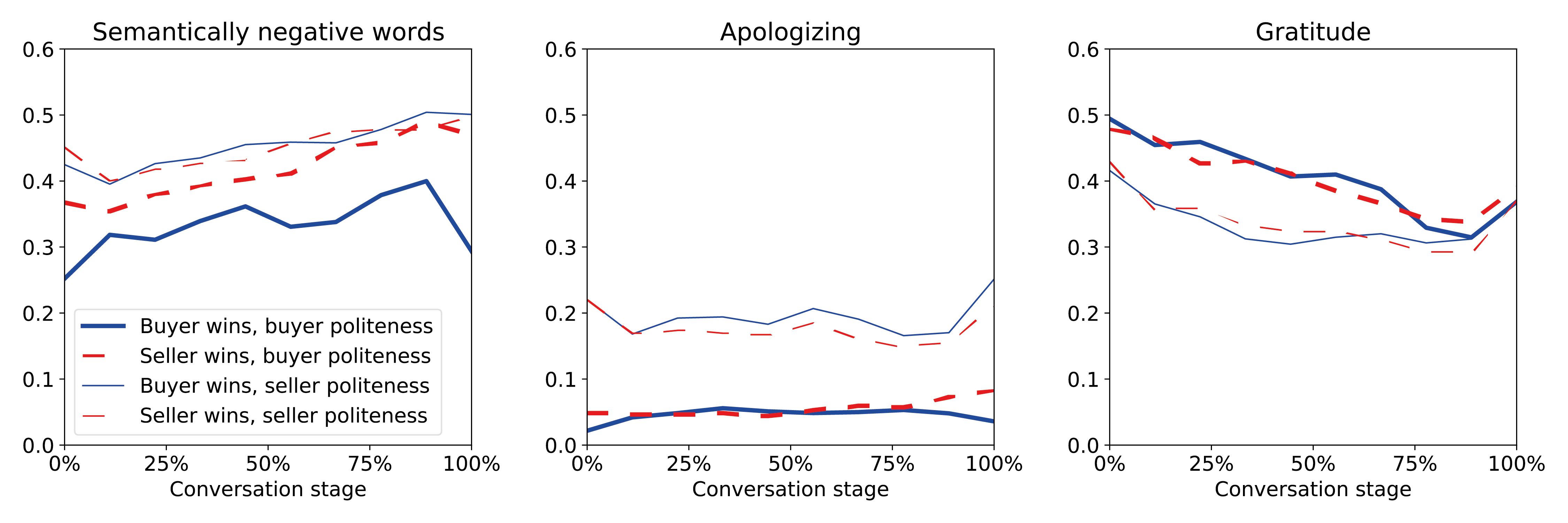}
\caption{\small Trajectory of negative words, apologies, and gratitude over the dispute. \label{fig:politeness} 
}
\end{figure*}

In addition to predicting the dispute outcome, we wished to examine dispute trajectory and its affect on buyers and sellers -- both during the dispute and after it ended. 

\subsubsection{During the Dispute}

To understand how the dispute affects the buyer and seller, we observe the textual messages they exchange. These messages enable glimpsing into their mood (e.g., annoyance or gratitude) and its evolution throughout the dispute. A particularly useful metric for our purpose is \emph{politeness}. 
Politeness is a method of communication that attempts to prevent the other party from being offended \cite{Brown}. By observing politeness of disputants, we can gain both access to their internal state, as well as insight on the effect of communication strategies on a participant's interests. 

\paragraph{Politeness Model} We extracted politeness features from the correspondence using the computational politeness model \cite{danescu2013computational}. In the model, politeness is divided into 21 strategies, such as \textit{greeting}, \textit{deference}, and \textit{apologizing}. Each strategy is a binary feature signifying whether it was used in a certain utterance of the conversation. 

First, we studied the effect of a politeness strategy in the first message of the buyer/seller on the outcome of the dispute. We collected 10,000 disputes with C2C sellers to avoid automated messages, and examined the first message to avoid effects that result from earlier stages of the conversation (e.g., buyer gratitude at the end of the conversation might just indicate the buyer won, and not that gratitude leads to buyer winning). 

When observing the correlation between the politeness strategies of the first message and the dispute outcome, we found that for almost all politeness strategies, showing any politeness strategy by the buyer tends to result in a worse outcome for them (Figure \ref{fig:first_polite}). This was true for both positive (polite) and negative (rude) politeness strategies. The only strategies that were significantly beneficial for the buyer were indirect greeting (e.g., ``Hey, I just wanted to...''), please opening (e.g., ``Please send me the...''), and indicative requests (e.g., ``Can you send me the...''). For the seller, the situation was different - most politeness strategies had no significant correlation with the outcome of the dispute, and the ones that did were generally weaker compared to the buyer correlations. 

To study how politeness evolves during conversation, we used a similar method to Danescu et al \cite{danescu2013no}, normalizing conversation length to test whether politeness changes over time (Figure \ref{fig:politeness}). In each trajectory, we separated seller and buyer messages, and tested their politeness trajectories conditioned on who won the dispute. We found that in all politeness strategies, throughout the dispute trajectory, it was better for the buyer to use as few politeness strategies as possible. Note that as before, this was true both for positive and negative politeness strategies. We show three of the politeness trajectories in Figure \ref{fig:politeness}, to discuss specific phenomena: of users that employed the semantically negative words strategy, which measures usage of words with negative sentiment (e.g. accuse, blame, complaint),
buyers who won used fewer negative words than either buyers that lost or sellers. Sellers were more prone to use the apologizing strategy than buyers, who rarely apologized throughout. Finally, in the gratitude strategy, buyers were more prone to offer gratitude in the beginning of the dispute, but as the dispute progressed both buyers and sellers showed diminishing gratitude, until (but not including) the final message of the conversation. 
\vspace{-2mm}

\subsubsection{After the Dispute} Disputes can be a disruptive event for buyers and sellers. Here we studied the effect of winning or losing a dispute on the future transactions of buyers (\emph{soft churn}).

To study the effect of participating in a dispute on buyers, we compared the number of transactions 7 weeks before and after the dispute over 532,552 buyers. To isolate the effect of the dispute, we chose periods of 15 weeks with only a single dispute in week 8. 
Participation in disputes had two effects: first, after an increase of activity prior to the dispute, there was a sharp decrease to a lower average number of transactions compared to before the dispute. Second, although the number of transactions was reduced for all buyers participating in a dispute, the ratio of post-dispute transactions and pre-dispute transactions was lower for buyers who lost (0.82) compared to the buyers who won the dispute (0.86). 

Finally, The percentage of buyers who did not purchase anything in the 7 weeks following a lost dispute was 33\% higher than the percentage of buyers who did not purchase anything in the 7 weeks following a won dispute (9\% vs 12\% respectively). Thus, we can see that buyers that lose a dispute tend to buy 
less afterwards, while buyers who win a dispute 
continue buying after the dispute ended.

\section{Conclusion and Future Work}
In this work we presented the first comprehensive study of dispute resolution in a large online marketplace. We present a model that assists human agents in resolving online disputes in \ecom. To this end, we developed a classifier with high accuracy, applied interpretability tools to explain the algorithm's decision to the human agent, and studied the effect of disputes on participants while they lasted and after they ended. While our model was developed based on data from one specific marketplace, the dispute process we described and the features we used, as well as the dispute claims (Item-not-received and significantly-not-as-described) and transaction types (B2C and C2C) represent common concepts that are applicable in many other online marketplaces.

Although our algorithm has succeeded in the prediction task, this work has several limitations. We have focused our research on cases with a clear outcome in favor of one side, and future efforts should expand to fuzzier labels that are presumably harder to classify. Due to privacy concerns we cannot publicly release the dataset, and could not evaluate our algorithm on similar datasets; however, we have tried to describe our data and algorithm in detail, to enable easy application to other datasets.

Several future directions could be interesting to explore: explanations of algorithmic classifications could be improved to better aid the arbitrator in the final call; incorporation of external rule-based knowledge source, such as decision trees, as part of the classification process; and deploying the model into live resolution systems, allowing integration with the decision making process. Such integration could help measure the time saved by the algorithm, and could lead to a further study on appeal rates. 

As  mentioned, recent studies show that AI models that aid in judicial and management decision making can unwittingly inherit the biases of the humans making those decisions. Studying the biases in ODR and whether or not they manifest in this model would be instrumental to improving the decision making process and making it more fair. 

\noindent {\bf Acknowledgments:} The authors would like to thank the reviewers for their insightful comments. This work was supported by the European Research Council (ERC) under the European Union's Horizon 2020 research and innovation programme (grant no. 852686, SIAM) and US National Science Foundation, US-Israel Binational Science Foundation (NSF-BSF) grant 2017741 (Shahaf).
\label{sec:dis}

\balance
\bibliographystyle{ACM-Reference-Format}
\bibliography{references} 


\begin{thebibliography}{35}


\ifx \showCODEN    \undefined \def \showCODEN     #1{\unskip}     \fi
\ifx \showDOI      \undefined \def \showDOI       #1{#1}\fi
\ifx \showISBNx    \undefined \def \showISBNx     #1{\unskip}     \fi
\ifx \showISBNxiii \undefined \def \showISBNxiii  #1{\unskip}     \fi
\ifx \showISSN     \undefined \def \showISSN      #1{\unskip}     \fi
\ifx \showLCCN     \undefined \def \showLCCN      #1{\unskip}     \fi
\ifx \shownote     \undefined \def \shownote      #1{#1}          \fi
\ifx \showarticletitle \undefined \def \showarticletitle #1{#1}   \fi
\ifx \showURL      \undefined \def \showURL       {\relax}        \fi
\providecommand\bibfield[2]{#2}
\providecommand\bibinfo[2]{#2}
\providecommand\natexlab[1]{#1}
\providecommand\showeprint[2][]{arXiv:#2}

\bibitem[\protect\citeauthoryear{Angwin, Larson, Mattu, and Kirchner}{Angwin
  et~al\mbox{.}}{2016}]%
        {angwin2016machine}
\bibfield{author}{\bibinfo{person}{Julia Angwin}, \bibinfo{person}{Jeff
  Larson}, \bibinfo{person}{Surya Mattu}, {and} \bibinfo{person}{Lauren
  Kirchner}.} \bibinfo{year}{2016}\natexlab{}.
\newblock \showarticletitle{Machine bias: There's software used across the
  country to predict future criminals}.
\newblock \bibinfo{journal}{\emph{ProPublica}}  \bibinfo{volume}{23}
  (\bibinfo{year}{2016}).
\newblock


\bibitem[\protect\citeauthoryear{Bojanowski, Grave, Joulin, and
  Mikolov}{Bojanowski et~al\mbox{.}}{2017}]%
        {bojanowski2017enriching}
\bibfield{author}{\bibinfo{person}{Piotr Bojanowski}, \bibinfo{person}{Edouard
  Grave}, \bibinfo{person}{Armand Joulin}, {and} \bibinfo{person}{Tomas
  Mikolov}.} \bibinfo{year}{2017}\natexlab{}.
\newblock \showarticletitle{Enriching word vectors with subword information}.
\newblock \bibinfo{journal}{\emph{Transactions of the Association for
  Computational Linguistics}}  \bibinfo{volume}{5} (\bibinfo{year}{2017}),
  \bibinfo{pages}{135--146}.
\newblock


\bibitem[\protect\citeauthoryear{Brennan, Dieterich, and Ehret}{Brennan
  et~al\mbox{.}}{2009}]%
        {brennan2009evaluating}
\bibfield{author}{\bibinfo{person}{Tim Brennan}, \bibinfo{person}{William
  Dieterich}, {and} \bibinfo{person}{Beate Ehret}.}
  \bibinfo{year}{2009}\natexlab{}.
\newblock \showarticletitle{Evaluating the predictive validity of the COMPAS
  risk and needs assessment system}.
\newblock \bibinfo{journal}{\emph{Criminal Justice and Behavior}}
  \bibinfo{volume}{36}, \bibinfo{number}{1} (\bibinfo{year}{2009}),
  \bibinfo{pages}{21--40}.
\newblock


\bibitem[\protect\citeauthoryear{Carmel, Uziel, Guy, Mass, and Roitman}{Carmel
  et~al\mbox{.}}{2012}]%
        {carmel12kl}
\bibfield{author}{\bibinfo{person}{David Carmel}, \bibinfo{person}{Erel Uziel},
  \bibinfo{person}{Ido Guy}, \bibinfo{person}{Yosi Mass}, {and}
  \bibinfo{person}{Haggai Roitman}.} \bibinfo{year}{2012}\natexlab{}.
\newblock \showarticletitle{Folksonomy-Based Term Extraction for Word Cloud
  Generation}.
\newblock \bibinfo{journal}{\emph{ACM Trans. Intell. Syst. Technol.}}
  \bibinfo{volume}{3}, \bibinfo{number}{4}, Article \bibinfo{articleno}{60}
  (\bibinfo{year}{2012}), \bibinfo{numpages}{20}~pages.
\newblock
\showISSN{2157-6904}


\bibitem[\protect\citeauthoryear{Carneiro, Novais, Andrade, Zeleznikow, and
  Neves}{Carneiro et~al\mbox{.}}{2014}]%
        {Carneiro}
\bibfield{author}{\bibinfo{person}{Davide Carneiro}, \bibinfo{person}{Paulo
  Novais}, \bibinfo{person}{Francisco Andrade}, \bibinfo{person}{John
  Zeleznikow}, {and} \bibinfo{person}{Jose Neves}.}
  \bibinfo{year}{2014}\natexlab{}.
\newblock \showarticletitle{Online dispute resolution: an artificial
  intelligence perspective}.
\newblock \bibinfo{journal}{\emph{Artificial Intelligence Review}}
  (\bibinfo{year}{2014}).
\newblock
\urldef\tempurl%
\url{https://doi.org/10.1007/s10462-011-9305-z}
\showDOI{\tempurl}


\bibitem[\protect\citeauthoryear{Chen and Guestrin}{Chen and Guestrin}{2016}]%
        {chen2016xgboost}
\bibfield{author}{\bibinfo{person}{Tianqi Chen} {and} \bibinfo{person}{Carlos
  Guestrin}.} \bibinfo{year}{2016}\natexlab{}.
\newblock \showarticletitle{Xgboost: A scalable tree boosting system}. In
  \bibinfo{booktitle}{\emph{Proceedings of the 22nd acm sigkdd international
  conference on knowledge discovery and data mining}}. ACM,
  \bibinfo{pages}{785--794}.
\newblock


\bibitem[\protect\citeauthoryear{Danescu-Niculescu-Mizil, Sudhof, Jurafsky,
  Leskovec, and Potts}{Danescu-Niculescu-Mizil et~al\mbox{.}}{2013a}]%
        {danescu2013computational}
\bibfield{author}{\bibinfo{person}{Cristian Danescu-Niculescu-Mizil},
  \bibinfo{person}{Moritz Sudhof}, \bibinfo{person}{Dan Jurafsky},
  \bibinfo{person}{Jure Leskovec}, {and} \bibinfo{person}{Christopher Potts}.}
  \bibinfo{year}{2013}\natexlab{a}.
\newblock \showarticletitle{A computational approach to politeness with
  application to social factors}. In \bibinfo{booktitle}{\emph{Proceedings of
  the 51st Annual Meeting of the Association for Computational Linguistics
  (Volume 1: Long Papers)}}. \bibinfo{pages}{250--259}.
\newblock


\bibitem[\protect\citeauthoryear{Danescu-Niculescu-Mizil, West, Jurafsky,
  Leskovec, and Potts}{Danescu-Niculescu-Mizil et~al\mbox{.}}{2013b}]%
        {danescu2013no}
\bibfield{author}{\bibinfo{person}{Cristian Danescu-Niculescu-Mizil},
  \bibinfo{person}{Robert West}, \bibinfo{person}{Dan Jurafsky},
  \bibinfo{person}{Jure Leskovec}, {and} \bibinfo{person}{Christopher Potts}.}
  \bibinfo{year}{2013}\natexlab{b}.
\newblock \showarticletitle{No country for old members: User lifecycle and
  linguistic change in online communities}. In
  \bibinfo{booktitle}{\emph{Proceedings of the 22nd international conference on
  World Wide Web}}. ACM, \bibinfo{pages}{307--318}.
\newblock


\bibitem[\protect\citeauthoryear{Dressel and Farid}{Dressel and Farid}{2018}]%
        {dressel2018accuracy}
\bibfield{author}{\bibinfo{person}{Julia Dressel} {and} \bibinfo{person}{Hany
  Farid}.} \bibinfo{year}{2018}\natexlab{}.
\newblock \showarticletitle{The accuracy, fairness, and limits of predicting
  recidivism}.
\newblock \bibinfo{journal}{\emph{Science advances}} \bibinfo{volume}{4},
  \bibinfo{number}{1} (\bibinfo{year}{2018}), \bibinfo{pages}{eaao5580}.
\newblock


\bibitem[\protect\citeauthoryear{Finale Doshi-Velez}{Finale
  Doshi-Velez}{2017}]%
        {kim}
\bibfield{author}{\bibinfo{person}{Been~Kim Finale Doshi-Velez}.}
  \bibinfo{year}{2017}\natexlab{}.
\newblock \showarticletitle{Towards A Rigorous Science of Interpretable Machine
  Learning}.
\newblock \bibinfo{journal}{\emph{arXiv preprint arXiv:1702.08608}}
  (\bibinfo{year}{2017}).
\newblock


\bibitem[\protect\citeauthoryear{Friedman, Anderson, Brett, Olekalns, Goates,
  and Lisco}{Friedman et~al\mbox{.}}{2004}]%
        {friedman}
\bibfield{author}{\bibinfo{person}{Ray Friedman}, \bibinfo{person}{Cameron
  Anderson}, \bibinfo{person}{Jeanne Brett}, \bibinfo{person}{Mara Olekalns},
  \bibinfo{person}{Nathan Goates}, {and} \bibinfo{person}{Cara~Cherry Lisco}.}
  \bibinfo{year}{2004}\natexlab{}.
\newblock \showarticletitle{The positive and negative effects of anger on
  dispute resolution: evidence from electronically mediated disputes}.
\newblock \bibinfo{journal}{\emph{Journal of Applied Psychology}}
  (\bibinfo{year}{2004}).
\newblock


\bibitem[\protect\citeauthoryear{Goodman}{Goodman}{2003}]%
        {goodman}
\bibfield{author}{\bibinfo{person}{Joseph~W. Goodman}.}
  \bibinfo{year}{2003}\natexlab{}.
\newblock \showarticletitle{The Pros and Cons of Online Dispute Resolution: An
  Assessment of Cyber-Mediation Websites}.
\newblock \bibinfo{journal}{\emph{Duke Law and Technology Review}}
  (\bibinfo{year}{2003}).
\newblock


\bibitem[\protect\citeauthoryear{Gunning}{Gunning}{2017}]%
        {gunning2017explainable}
\bibfield{author}{\bibinfo{person}{David Gunning}.}
  \bibinfo{year}{2017}\natexlab{}.
\newblock \showarticletitle{Explainable artificial intelligence (xai)}.
\newblock \bibinfo{journal}{\emph{Defense Advanced Research Projects Agency
  (DARPA), nd Web}}  \bibinfo{volume}{2} (\bibinfo{year}{2017}).
\newblock


\bibitem[\protect\citeauthoryear{Guy}{Guy}{2018}]%
        {recsys}
\bibfield{author}{\bibinfo{person}{Ido Guy}.} \bibinfo{year}{2018}\natexlab{}.
\newblock \showarticletitle{Connecting Sellers and Buyers on the World's
  Largest Inventory}. In \bibinfo{booktitle}{\emph{Proc. of RecSys}}.
  \bibinfo{pages}{490--491}.
\newblock


\bibitem[\protect\citeauthoryear{Joulin, Grave, Bojanowski, and Mikolov}{Joulin
  et~al\mbox{.}}{2016}]%
        {joulin2016bag}
\bibfield{author}{\bibinfo{person}{Armand Joulin}, \bibinfo{person}{Edouard
  Grave}, \bibinfo{person}{Piotr Bojanowski}, {and} \bibinfo{person}{Tomas
  Mikolov}.} \bibinfo{year}{2016}\natexlab{}.
\newblock \showarticletitle{Bag of Tricks for Efficient Text Classification}.
\newblock \bibinfo{journal}{\emph{arXiv preprint arXiv:1607.01759}}
  (\bibinfo{year}{2016}).
\newblock


\bibitem[\protect\citeauthoryear{Katsh, Rifkin, and Gaitenby}{Katsh
  et~al\mbox{.}}{1999}]%
        {katsh1999commerce}
\bibfield{author}{\bibinfo{person}{Ethan Katsh}, \bibinfo{person}{Janet
  Rifkin}, {and} \bibinfo{person}{Alan Gaitenby}.}
  \bibinfo{year}{1999}\natexlab{}.
\newblock \showarticletitle{E-Commerce, E-Disputes, and E-Dispute Resolution:
  in the shadow of eBay law}.
\newblock \bibinfo{journal}{\emph{Ohio St. J. on Disp. Resol.}}
  \bibinfo{volume}{15} (\bibinfo{year}{1999}), \bibinfo{pages}{705}.
\newblock


\bibitem[\protect\citeauthoryear{Kleinberg, Lakkaraju, Leskovec, Ludwig, and
  Mullainathan}{Kleinberg et~al\mbox{.}}{2017}]%
        {Kleinberg}
\bibfield{author}{\bibinfo{person}{Jon Kleinberg}, \bibinfo{person}{Himabindu
  Lakkaraju}, \bibinfo{person}{Jure Leskovec}, \bibinfo{person}{Jens Ludwig},
  {and} \bibinfo{person}{Sendhil Mullainathan}.}
  \bibinfo{year}{2017}\natexlab{}.
\newblock \showarticletitle{{Human Decisions and Machine Predictions*}}.
\newblock \bibinfo{journal}{\emph{The Quarterly Journal of Economics}}
  \bibinfo{volume}{133}, \bibinfo{number}{1} (\bibinfo{date}{08}
  \bibinfo{year}{2017}), \bibinfo{pages}{237--293}.
\newblock
\showISSN{0033-5533}
\urldef\tempurl%
\url{https://doi.org/10.1093/qje/qjx032}
\showDOI{\tempurl}
\showeprint{http://oup.prod.sis.lan/qje/article-pdf/133/1/237/24246094/qjx032.pdf}


\bibitem[\protect\citeauthoryear{Korobov and Lopuhin}{Korobov and
  Lopuhin}{2020}]%
        {ELI5}
\bibfield{author}{\bibinfo{person}{Mikhail Korobov} {and}
  \bibinfo{person}{Konstantin Lopuhin}.} \bibinfo{year}{2020}\natexlab{}.
\newblock \bibinfo{title}{ELI5}.
\newblock \bibinfo{howpublished}{\url{https://eli5.readthedocs.io}}.
\newblock


\bibitem[\protect\citeauthoryear{Lach and Economou}{Lach and Economou}{2019}]%
        {fourpr}
\bibfield{author}{\bibinfo{person}{Eileen~M. Lach} {and}
  \bibinfo{person}{Nicolas Economou}.} \bibinfo{year}{2019}\natexlab{}.
\newblock \showarticletitle{Four Principles for the Trustworthy Adoption of AI
  in Legal Systems}.
\newblock \bibinfo{journal}{\emph{Bloomberg Law}} (\bibinfo{year}{2019}).
\newblock


\bibitem[\protect\citeauthoryear{Lawlor}{Lawlor}{1963}]%
        {lawlor}
\bibfield{author}{\bibinfo{person}{Reed~C. Lawlor}.}
  \bibinfo{year}{1963}\natexlab{}.
\newblock \showarticletitle{What Computers Can Do: Analysis and Prediction of
  Judicial Decisions}.
\newblock \bibinfo{journal}{\emph{American Bar Association Journal}}
  (\bibinfo{date}{4} \bibinfo{year}{1963}).
\newblock


\bibitem[\protect\citeauthoryear{Lu~Wang}{Lu~Wang}{2014}]%
        {wang}
\bibfield{author}{\bibinfo{person}{Claire~Cardie Lu~Wang}.}
  \bibinfo{year}{2014}\natexlab{}.
\newblock \showarticletitle{A Piece of My Mind: A Sentiment Analysis Approach
  for Online Dispute Detection}.
\newblock \bibinfo{journal}{\emph{Annual Conference of the Association for
  Computational Linguistics}} (\bibinfo{date}{6} \bibinfo{year}{2014}).
\newblock


\bibitem[\protect\citeauthoryear{Lundberg and Lee}{Lundberg and Lee}{2017}]%
        {NIPS2017_7062}
\bibfield{author}{\bibinfo{person}{Scott~M Lundberg} {and}
  \bibinfo{person}{Su-In Lee}.} \bibinfo{year}{2017}\natexlab{}.
\newblock \showarticletitle{A Unified Approach to Interpreting Model
  Predictions}.
\newblock In \bibinfo{booktitle}{\emph{Advances in Neural Information
  Processing Systems 30}}, \bibfield{editor}{\bibinfo{person}{I.~Guyon},
  \bibinfo{person}{U.~V. Luxburg}, \bibinfo{person}{S.~Bengio},
  \bibinfo{person}{H.~Wallach}, \bibinfo{person}{R.~Fergus},
  \bibinfo{person}{S.~Vishwanathan}, {and} \bibinfo{person}{R.~Garnett}}
  (Eds.). \bibinfo{publisher}{Curran Associates, Inc.},
  \bibinfo{pages}{4765--4774}.
\newblock
\urldef\tempurl%
\url{http://papers.nips.cc/paper/7062-a-unified-approach-to-interpreting-model-predictions.pdf}
\showURL{%
\tempurl}


\bibitem[\protect\citeauthoryear{Molnar}{Molnar}{2019}]%
        {molnar}
\bibfield{author}{\bibinfo{person}{Christoph Molnar}.}
  \bibinfo{year}{2019}\natexlab{}.
\newblock \bibinfo{booktitle}{\emph{Interpretable machine learning}}.
\newblock \bibinfo{publisher}{Lulu.com}.
\newblock
\urldef\tempurl%
\url{https://christophm.github.io/interpretable-ml-book/}
\showURL{%
\tempurl}


\bibitem[\protect\citeauthoryear{Monahan and Skeem}{Monahan and Skeem}{2016}]%
        {monahan2016risk}
\bibfield{author}{\bibinfo{person}{John Monahan} {and}
  \bibinfo{person}{Jennifer~L Skeem}.} \bibinfo{year}{2016}\natexlab{}.
\newblock \showarticletitle{Risk assessment in criminal sentencing}.
\newblock \bibinfo{journal}{\emph{Annual review of clinical psychology}}
  \bibinfo{volume}{12} (\bibinfo{year}{2016}), \bibinfo{pages}{489--513}.
\newblock


\bibitem[\protect\citeauthoryear{Morison and Harkens}{Morison and
  Harkens}{2019}]%
        {Morison}
\bibfield{author}{\bibinfo{person}{John Morison} {and} \bibinfo{person}{Adam
  Harkens}.} \bibinfo{year}{2019}\natexlab{}.
\newblock \showarticletitle{Re-engineering justice? Robot judges, computerised
  courts and (semi) automated legal decision-making}.
\newblock \bibinfo{journal}{\emph{Legal Studies}} (\bibinfo{date}{12}
  \bibinfo{year}{2019}).
\newblock
\urldef\tempurl%
\url{https://doi.org/10.1017/lst.2019.5}
\showDOI{\tempurl}


\bibitem[\protect\citeauthoryear{Pedregosa, Varoquaux, Gramfort, Michel,
  Thirion, Grisel, Blondel, Prettenhofer, Weiss, Dubourg, Vanderplas, Passos,
  Cournapeau, Brucher, Perrot, and Duchesnay}{Pedregosa et~al\mbox{.}}{2011}]%
        {scikit-learn}
\bibfield{author}{\bibinfo{person}{F. Pedregosa}, \bibinfo{person}{G.
  Varoquaux}, \bibinfo{person}{A. Gramfort}, \bibinfo{person}{V. Michel},
  \bibinfo{person}{B. Thirion}, \bibinfo{person}{O. Grisel},
  \bibinfo{person}{M. Blondel}, \bibinfo{person}{P. Prettenhofer},
  \bibinfo{person}{R. Weiss}, \bibinfo{person}{V. Dubourg}, \bibinfo{person}{J.
  Vanderplas}, \bibinfo{person}{A. Passos}, \bibinfo{person}{D. Cournapeau},
  \bibinfo{person}{M. Brucher}, \bibinfo{person}{M. Perrot}, {and}
  \bibinfo{person}{E. Duchesnay}.} \bibinfo{year}{2011}\natexlab{}.
\newblock \showarticletitle{Scikit-learn: Machine Learning in {P}ython}.
\newblock \bibinfo{journal}{\emph{Journal of Machine Learning Research}}
  \bibinfo{volume}{12} (\bibinfo{year}{2011}), \bibinfo{pages}{2825--2830}.
\newblock


\bibitem[\protect\citeauthoryear{Penelope~Brown}{Penelope~Brown}{1987}]%
        {Brown}
\bibfield{author}{\bibinfo{person}{Stephen C.~Levinson Penelope~Brown}.}
  \bibinfo{year}{1987}\natexlab{}.
\newblock \bibinfo{booktitle}{\emph{Politeness: Some Universals in Language
  Usage}}.
\newblock \bibinfo{publisher}{Cambridge university press}.
\newblock


\bibitem[\protect\citeauthoryear{Ribeiro, Singh, and Guestrin}{Ribeiro
  et~al\mbox{.}}{2016}]%
        {ribeiro2016should}
\bibfield{author}{\bibinfo{person}{Marco~Tulio Ribeiro},
  \bibinfo{person}{Sameer Singh}, {and} \bibinfo{person}{Carlos Guestrin}.}
  \bibinfo{year}{2016}\natexlab{}.
\newblock \showarticletitle{Why should i trust you?: Explaining the predictions
  of any classifier}. In \bibinfo{booktitle}{\emph{Proceedings of the 22nd ACM
  SIGKDD international conference on knowledge discovery and data mining}}.
  ACM, \bibinfo{pages}{1135--1144}.
\newblock


\bibitem[\protect\citeauthoryear{Rossi}{Rossi}{2019}]%
        {rossi2019building}
\bibfield{author}{\bibinfo{person}{Francesca Rossi}.}
  \bibinfo{year}{2019}\natexlab{}.
\newblock \showarticletitle{Building trust in artificial intelligence}.
\newblock \bibinfo{journal}{\emph{Journal of international affairs}}
  \bibinfo{volume}{72}, \bibinfo{number}{1} (\bibinfo{year}{2019}),
  \bibinfo{pages}{127--134}.
\newblock


\bibitem[\protect\citeauthoryear{Rule}{Rule}{2016}]%
        {rule2016designing}
\bibfield{author}{\bibinfo{person}{Colin Rule}.}
  \bibinfo{year}{2016}\natexlab{}.
\newblock \showarticletitle{Designing a Global Online Dispute Resolution
  System: Lessons Learned from eBay}.
\newblock \bibinfo{journal}{\emph{U. St. Thomas LJ}}  \bibinfo{volume}{13}
  (\bibinfo{year}{2016}), \bibinfo{pages}{354}.
\newblock


\bibitem[\protect\citeauthoryear{Sela}{Sela}{2012}]%
        {sela}
\bibfield{author}{\bibinfo{person}{Ayelet Sela}.}
  \bibinfo{year}{2012}\natexlab{}.
\newblock \showarticletitle{Can Computers Be Fair? How Automated and
  Human-Powered Online Dispute Resolution Affect Procedural Justice in
  Mediation and Arbitration}.
\newblock \bibinfo{journal}{\emph{Ohio State Journal on Dispute Resolution}}
  (\bibinfo{date}{1} \bibinfo{year}{2012}).
\newblock


\bibitem[\protect\citeauthoryear{Steelman}{Steelman}{1997}]%
        {steelman1997have}
\bibfield{author}{\bibinfo{person}{David~C Steelman}.}
  \bibinfo{year}{1997}\natexlab{}.
\newblock \showarticletitle{What Have We Learned About Court Delay, "Local
  Legal Culture," and Caseflow Management Since the Late 1970s?}
\newblock \bibinfo{journal}{\emph{Justice System Journal}}
  \bibinfo{volume}{19}, \bibinfo{number}{2} (\bibinfo{year}{1997}),
  \bibinfo{pages}{145--166}.
\newblock


\bibitem[\protect\citeauthoryear{Xu, Smith, Murray, and Woolf}{Xu
  et~al\mbox{.}}{2012}]%
        {xu2012analyzing}
\bibfield{author}{\bibinfo{person}{Xiaoxi Xu}, \bibinfo{person}{David Smith},
  \bibinfo{person}{Tom Murray}, {and} \bibinfo{person}{B Woolf}.}
  \bibinfo{year}{2012}\natexlab{}.
\newblock \showarticletitle{Analyzing Conflict Narratives to Predict
  Settlements in EBay Feedback Dispute Resolution}. In
  \bibinfo{booktitle}{\emph{Proceedings of the 2012 International Conference on
  Data Mining (DMIN 2012), Las Vegas (July 2012)}}.
\newblock


\bibitem[\protect\citeauthoryear{Zeleznikow}{Zeleznikow}{2017}]%
        {Zeleznikow}
\bibfield{author}{\bibinfo{person}{John Zeleznikow}.}
  \bibinfo{year}{2017}\natexlab{}.
\newblock \showarticletitle{Can Artificial Intelligence and Online Dispute
  Resolution Enhance Efficiency and Effectiveness in Courts}.
\newblock \bibinfo{journal}{\emph{International Journal for Court
  Administration}} (\bibinfo{date}{05} \bibinfo{year}{2017}).
\newblock


\bibitem[\protect\citeauthoryear{Zhou, Zhang, Liu, Sun, and Si}{Zhou
  et~al\mbox{.}}{2019}]%
        {zhou2019legal}
\bibfield{author}{\bibinfo{person}{Xin Zhou}, \bibinfo{person}{Yating Zhang},
  \bibinfo{person}{Xiaozhong Liu}, \bibinfo{person}{Changlong Sun}, {and}
  \bibinfo{person}{Luo Si}.} \bibinfo{year}{2019}\natexlab{}.
\newblock \showarticletitle{Legal Intelligence for E-commerce: Multi-task
  Learning by Leveraging Multiview Dispute Representation}. In
  \bibinfo{booktitle}{\emph{Proceedings of the 42nd International ACM SIGIR
  Conference on Research and Development in Information Retrieval}}.
  \bibinfo{pages}{315--324}.
\newblock


\end{thebibliography}

\end{document}